\pdfoutput=1
\documentclass[journal]{IEEEtran}
\usepackage{graphicx}
\usepackage{array}
\usepackage{multicol}
\usepackage{xcolor} 
\usepackage{caption}
\usepackage{float} 
\usepackage{svg}
\usepackage{booktabs}

\usepackage{enumitem}
\usepackage{amsmath}
\usepackage{booktabs}
\usepackage{makecell}
\usepackage{graphicx}
\usepackage{amssymb}
\usepackage{multirow}
\usepackage{bbding}
\usepackage[justification=centering]{caption}
\usepackage{ulem}
\usepackage[square,comma,sort&compress]{natbib}
\setcitestyle{numbers,square}
\usepackage{hyperref}
\hypersetup{hypertex=true,colorlinks=true,linkcolor=blue,anchorcolor=blue,citecolor=blue}
\usepackage{cite}
\usepackage{subfigure} 

\usepackage{hyperref}
\hypersetup{
    colorlinks=true,  
    linkcolor=black,  
    urlcolor=black  
}

\usepackage{caption}
\ifCLASSINFOpdf
\else
\fi

\begin{document}
\title{MEET: A Million-Scale Dataset for Fine-Grained Geospatial Scene Classification with Zoom-Free Remote Sensing Imagery}

\author{Yansheng Li,~\IEEEmembership{Senior Member,~IEEE,}
 Yuning Wu,
Gong Cheng,~\IEEEmembership{Member,~IEEE,}
 Chao Tao,
 Bo Dang,
 
 Yu Wang,
 Jiahao Zhang,
 Chuge Zhang,
  Yiting Liu,
 Xu Tang,~\IEEEmembership{Senior Member,~IEEE,}

 Jiayi Ma,~\IEEEmembership{Senior Member,~IEEE,}
        and~Yongjun Zhang,~\IEEEmembership{Member,~IEEE}

}


\maketitle
\begin{abstract}
Accurate fine-grained geospatial scene classification using remote sensing imagery is essential for a wide range of applications. However, existing approaches often rely on manually zooming remote sensing images at different scales to create typical scene samples. This approach fails to adequately support the fixed-resolution image interpretation requirements in real-world scenarios.
To address this limitation, we introduce the Million-scale finE-grained geospatial scEne classification dataseT (MEET), which contains over 1.03 million zoom-free remote sensing scene samples, manually annotated into 80 fine-grained categories. In MEET, each scene sample follows a scene-in-scene layout, where the central scene serves as the reference, and auxiliary scenes provide crucial spatial context for fine-grained classification. 
Moreover, to tackle the emerging challenge of scene-in-scene classification, we present the Context-Aware Transformer (CAT), a model specifically designed for this task, which adaptively fuses spatial context to accurately classify the scene samples. CAT adaptively fuses spatial context to accurately classify the scene samples by learning attentional features that capture the relationships between the center and auxiliary scenes.
Based on MEET, we establish a comprehensive benchmark for fine-grained geospatial scene classification, evaluating CAT against 11 competitive baselines. The results demonstrate that CAT significantly outperforms these baselines, achieving a 1.88\% higher balanced accuracy (BA) with the Swin-Large backbone, and a notable 7.87\% improvement with the Swin-Huge backbone. Further experiments validate the effectiveness of each module in CAT and show the practical applicability of CAT in the urban functional zone mapping.
The source code and dataset will be publicly available at https://jerrywyn.github.io/project/MEET.html.
\end{abstract}

\begin{IEEEkeywords}
Fine-grained geospatial scene classification, remote sensing imagery, transformer, million-scale dataset, scene-in-scene.
\end{IEEEkeywords}

\IEEEpeerreviewmaketitle

\begin{figure}[htp]
    \centering  

    \subfigure[Samples from the existing FGSC dataset (e.g., NWPU)]{
        \includegraphics[width=0.42\textwidth]
        {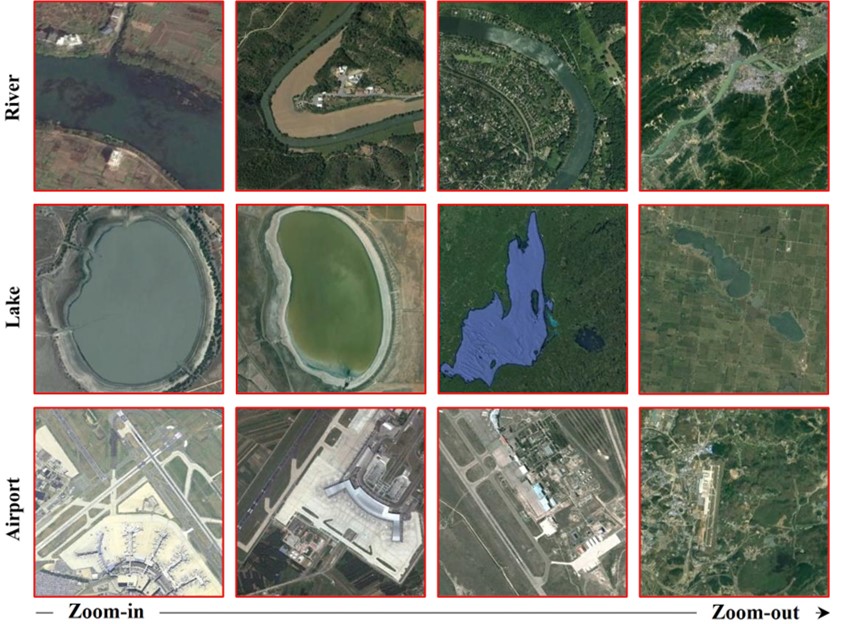}
        \label{fig:vary}
    }
    \hspace*{-0.32cm}
    \subfigure[Samples from our MEET dataset]{
        \includegraphics[width=0.42\textwidth]
        {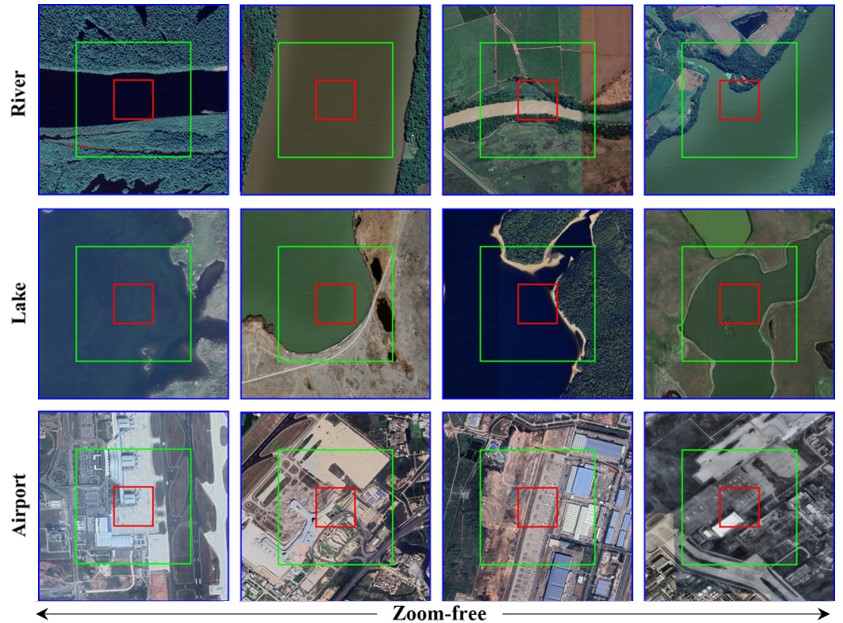}
        \label{fig:fix}
    }

        \captionsetup{justification=justified, singlelinecheck=false}    
        \caption{Illustration of the zoom-free and fine-grained characteristics of our MEET dataset.
Fig. 1(a) shows that the existing FGSC dataset forms typical scene samples by manually zooming remote sensing images at different rates.
In Fig. 1(b), the center scene outlined in \textcolor{red}{red}, is the basic unit for classification, while the surrounding scene outlined in \textcolor{green}{green} and global scene outlined in \textcolor{blue}{blue} serve as auxiliary contextual images.
With zoom-free samples and auxiliary scenes, MEET addresses inter-class and intra-class confusion in zoom-free image samples.}
    \label{fig:intro00}
\end{figure}

\section{Introduction}
\IEEEPARstart{F}{ine-grained} geospatial scene classification (FGSC) with remote sensing imagery (RSI) aims to categorize scene samples into fine-grained geospatial scene categories. 
Compared to coarse-grained remote sensing scene classification~\citep{xie2019scale,tang2021attention,sun2019remote,zou2015deep}, FGSC presents greater challenges but offers significant utility in various applications, such as water resource management~\citep{phinn2012multi,mishra2014mapping}, urban planning~\citep{yang2020small}, habitat conservation~\citep{gamba2012human,martha2011segment,cheng2013automatic}, et al.
Along with increasing high-resolution RSI~\citep{fu2017classification,tong2020land} and powerful deep learning models~\citep{liu2022swin,guo2024skysense} available, 
FGSC holds substantial promise and has become a focal point of research~\citep{long2021creating}.

\begin{figure}[t]
    \centering  

\includegraphics[width=0.49\textwidth]{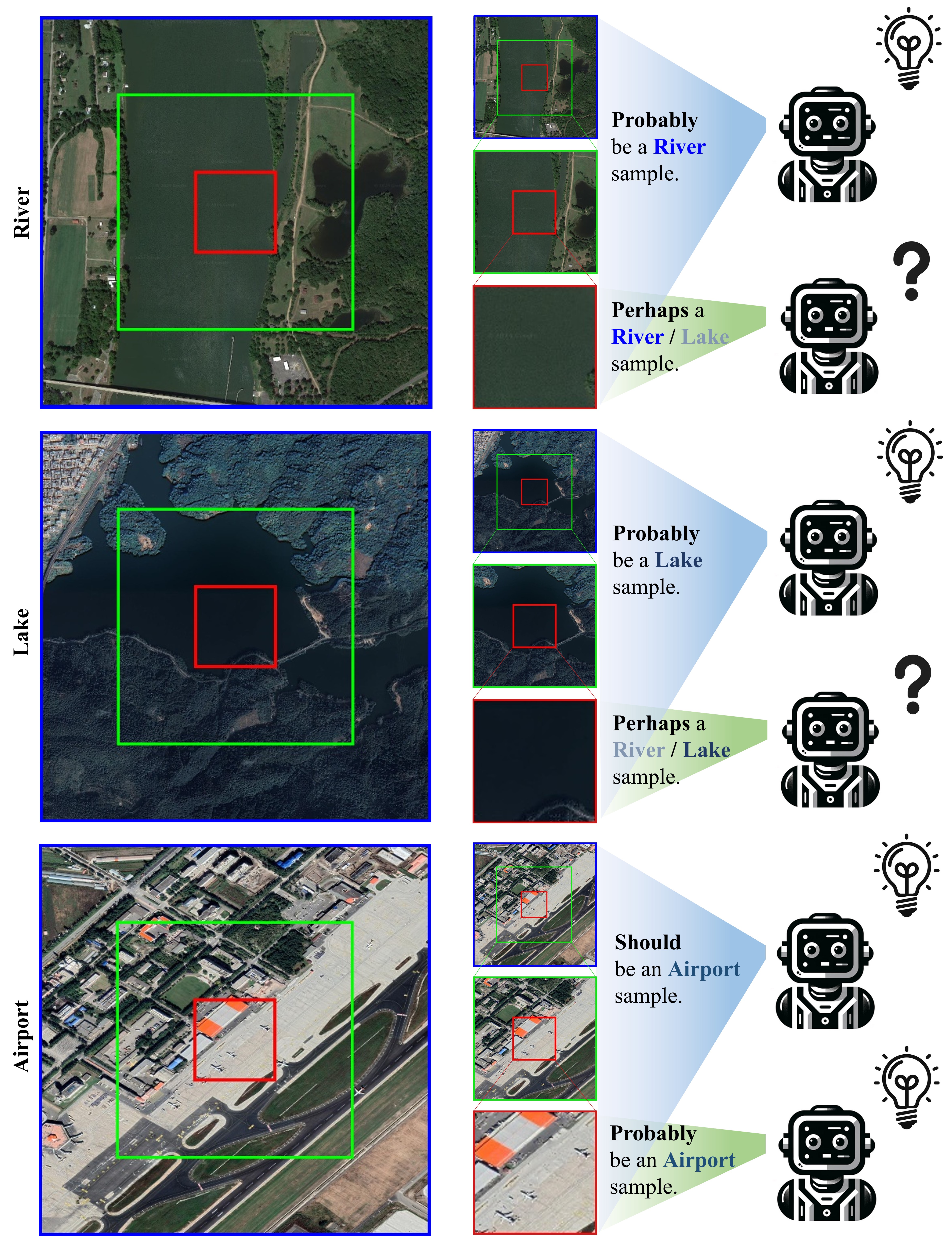}
        \captionsetup{justification=justified, singlelinecheck=false}    \caption{ Superiority illustration of the remote sensing image scene sample with the scene-in-scene layout. The sample labeled as an airport shows the case that models can successfully infer the fine-grained scene category using only the center scene and the auxiliary scenes may benefit improving the classification performance. 
        For the samples from first row and second row, models fail to predict the fine-grained scene category using only the center scene but has a great potential to obtain the right fine-grained scene category using both the center scene and the auxiliary scenes.}
    \label{fig:galaxy}
\end{figure}

To pursue FGSC, existing research~\citep{cheng2017remote,xia2017aid,long2021creating} manually zoom RSI with different rates to form typical scene samples (e.g., the samples from Fig.~\ref{fig:vary} ), which are further utilized to train FGSC models. However, in practical applications~\citep{hu2013tile,li2017unsupervised,huang2023comprehensive,xiao2023contextual,lu2022unified},
the input RSI to be classified is often with fixed-resolution. In this situation, the trained FGSC model with zooming samples may perform poorly. Without no doubt, manual zooming intervention of the input imagery may improve the scene classification performance but inevitably harm automatic process. With this consideration, this paper tries to leverage the fixed-resolution RSI without zooming to form scene samples. However, the zoom-free solution may raise another issue (e.g., the samples from Fig.~\ref{fig:fix} ) that fixed-resolution samples presents inter-class and intra-class confusion, especially for fine-grained categories.
To address this issue, we introduce scene-in-scene layout to form the scene sample. 
As shown in Fig.~\ref{fig:galaxy}, the center scene is the basic unit for classification, while the surrounding scene and global scene serve as auxiliary contexts.
Only focusing on center scene leads to confusion between river and lake categories, while this issue can be addressed by introducing auxiliary scenes.
In summary, as shown in Table~\ref{fig:all_dataset}, existing scene classification datasets have the following issues:
\textbf{(i)} manually zooming scene samples with different rates results in a gap with fixed-resulotion image interpretation in real-world application;
\textbf{(ii)} insufficient sample quantity and limited category coverage restrict the effectiveness of the dataset.

To address the above challenges, we introduce a Million-scale finE-grained geospatial scEne classification dataseT (\textbf{MEET}). 
In terms of sample size and category coverage, MEET contains 1.03 million samples across 80 geospatial scene categories, surpassing existing datasets~\citep{yang2010bag,xia2010structural,zou2015deep,basu2015deepsat,penatti2015deep,zhao2016feature,zhu2016bag,cheng2017remote,xia2017aid,xiao2017high,helber2019eurosat,zhou2018patternnet,wang2018scene,sumbul2019bigearthnet,li2020rsi,qi2020mlrsnet,li2020clrs,long2021creating,li2021robust,hua2021multiscene,yuan2022wh} in sample quantity, diversity, and fine granularity. This initiative is designed to provide a robust foundation for advancing scene classification methods and enhancing practical land-cover applications.
To address the challenge of avoiding zoomed scene samples, we incorporate surrounding and global scenes as essential auxiliary context, enriching the classification process. Each sample includes the image to be classified, along with two ranges of surrounding images captured from different field-of-views. Grouping these surrounding and global scenes together provides the necessary contextual information for accurate classification of the center scene. Furthermore, this organization offers scalability, enabling adaptive context fusion for varying classification tasks that require different ranges of context.

As illustrated in Fig.~\ref{fig:galaxy}, the center scene serves as the core unit for classification, but focusing solely on it can lead to confusion between categories, such as rivers and lakes. By incorporating auxiliary scenes, this issue can be resolved. The need for additional context depends on the specific classification task. For instance, in the airport category, the model can easily identify the salient object (i.e., an airplane) from the center scene, so no additional context is necessary. However, for categories like rivers and lakes, the introduction of auxiliary scenes becomes essential. These scenes provide a broader field-of-view, allowing the model to better discern features such as riverbanks and water bodies, which helps reduce both inter-class and intra-class confusion.

\begin{table*}[t]
\centering
\caption{ \\COMPARISON AMONG OPEN-SOURCE RSI SCENE CLASSIFICATION DATASETS
\\ Fine-grained characteristic refers to a dataset that contains more than 30 categories.
\\ * indicates the statistic of the publicly released part from the whole dataset.}
\begin{tabular}{lccccccc}
\toprule

Dataset & \makecell{Number of \\ categories} & \makecell{Number of \\ samples} & Spatial resolution (m) & Image size & Fine-grained & Zoom-free

\\
\midrule
UC-Merced~\citep{yang2010bag} & 21 & 2,100 & 0.3 & 256$\times$256 & × & \Checkmark 
\\
WHU-RS19~\citep{xia2010structural} & 19 & 1,013 & up to 0.5 & 600$\times$600 & × & ×  
\\
RSSCN7~\citep{zou2015deep} & 7 & 2,800 & -- & 400$\times$400 & × & --  
\\

SAT-4~\citep{basu2015deepsat} & 4 & 500,000 & 1-6 & 28$\times$28 & × & ×  
\\
SAT-6~\citep{basu2015deepsat} & 6 & 405,000 & 1-6 & 28$\times$28 & × & ×  
\\
BCS~\citep{penatti2015deep} & 2 & 2,876 & -- & -- & × & --  
\\
RSC11~\citep{zhao2016feature} & 11 & 1,232 & 0.2 & 512$\times$512 & × & \Checkmark 
\\
SIRI-WHU~\citep{zhu2016bag} & 12 & 2,400 & 2 & 200$\times$200 & × & \Checkmark  
\\

NWPU~\citep{cheng2017remote} & 45 & 31,500 & 0.2-30 & 256$\times$256 & \Checkmark & ×  
\\

AID~\citep{xia2017aid} & 30 & 10,000 & 0.5-8 & 600$\times$600 & \Checkmark & ×   
\\

RSD46-WHU~\citep{xiao2017high} & 46 & 117,000 & 0.5-2 & 256$\times$256 & \Checkmark & ×    
\\
EuroSAT~\citep{helber2019eurosat} & 10 & 27,000 & 10 & 64$\times$64 & × & \Checkmark  
\\
PatternNet~\citep{zhou2018patternnet} & 38 & 30,400 & 0.06-4.7 & 256$\times$256 & \Checkmark & ×  
\\
OPTIMAL-31~\citep{wang2018scene} & 31 & 1,860 & -- & 256$\times$256 & \Checkmark & --  
\\

BigEarthNet~\citep{sumbul2019bigearthnet} & 43 & 590,326 & 10,20,60 & [20$\times$20,120$\times$120] & \Checkmark & ×  
\\
RSI-CB256~\citep{li2020rsi} & 35 & 24,000 & 0.3-3 & 256$\times$256 & \Checkmark & ×  
\\
RSI-CB128~\citep{li2020rsi} & 45 & 36,000 & 0.3-3 & 128$\times$128 & \Checkmark & ×  
\\
MLRSN~\citep{qi2020mlrsnet} & 46 & 109,161 & 0.1-10 & 256$\times$256 & \Checkmark & ×  
\\
CLRS~\citep{li2020clrs} & 25 & 15,000 & 0.26-8.85 & 256$\times$256 & × & ×  
\\
Million-AID*~\citep{long2021creating} & 51 & 10,000 & 0.5-153 & [256$\times$256,512$\times$512] & \Checkmark & ×  
\\
SR-RSKG~\citep{li2021robust} & 70 & 56,000 & -- & 256$\times$256 & \Checkmark & × 
\\
Multiscene~\citep{hua2021multiscene} & 36 & 100,000 & 0.3-0.6 & 512$\times$512 & \Checkmark & × 
\\
WH-MAVS~\citep{yuan2022wh} & 14 & 47,137 & 1.2 & 200$\times$200 & × & \Checkmark 
 
  \\
\midrule
Our MEET & 80 & 1,033,778 & 1   &  
\{256$\times$ 256, 768$\times$ 768, 1280$\times$1280\}
& \Checkmark & \Checkmark
\\
\bottomrule
\end{tabular}

\label{fig:all_dataset}

\end{table*}

To advance the research on fundamental and practical issues, we propose a new challenging yet meaningful task: FGSC with zoom-free RSI. By contrast to existing FGSC methods~\citep{fang2022spatial,minetto2019hydra,wittich2018recommending} that used expensive external modalities to improve the interpretation of fine-grained categories, utilizing readily available surrounding RSIs is a more suitable and natural choice. Other potential difficulties can be categorized into two main aspects:
\textbf{(i)} avoiding performance degradation in most cases when integrating contextual information;
\textbf{(ii)} mitigating drastic memory consumption when applying contextual information, which limits practical usability.
Based on MEET, we propose a context-aware transformer (CAT) which can flexibly exploit RSI with multiple field-of-views for FGSC. CAT offers two key advantages: \textbf{(i)} leveraging spatial context information while avoiding performance degradation; \textbf{(ii)} being user-friendly, lightweight and compatible with pre-trained models. 
In summary, the MEET dataset and CAT framework proposed in this paper aim to establish a new benchmark for FGSC with zoom-free RSI. With the help of this benchmark, more innovative algorithms can be developed to facilitate the development of FGSC. 
The main contributions of this paper are as follows:

\begin{itemize}
\item We introduce MEET, the first million-scale dataset for FGSC with zoom-free RSI. It provides over 1.03 million samples where each sample employs a scene-in-scene layout, offering a new data organization for FGSC.
\item To avoid excessive memory consumption, a new CAT is proposed to address FGSC with zoom-free RSI and achieves progressive visual feature extraction through multi-level supervision.
\item We establish a new benchmark for FGSC with zoom-free RSI based on MEET. Comparisons with existing state-of-the-art algorithms demonstrate the superiority of our CAT. This benchmark may contribute to the fundamental evaluation of FGSC and promote the advancement of practical land-cover applications.

\end{itemize}

The rest of this paper is organized as follows: Section~\ref{sec:related_work} reviews existing FGSC datasets and algorithms. Section~\ref{sec:glh_dataset} describes the proposed MEET dataset in detail. Section~\ref{sec:proposed_method} introduces the proposed CAT for FGSC. Section~\ref{sec:experiment} presents the experimental results. Finally, Section~\ref{sec:conclusion} summarizes the paper and provides insights for future work.

\begin{figure*}[t]
\centering
	\includegraphics[scale=.54]{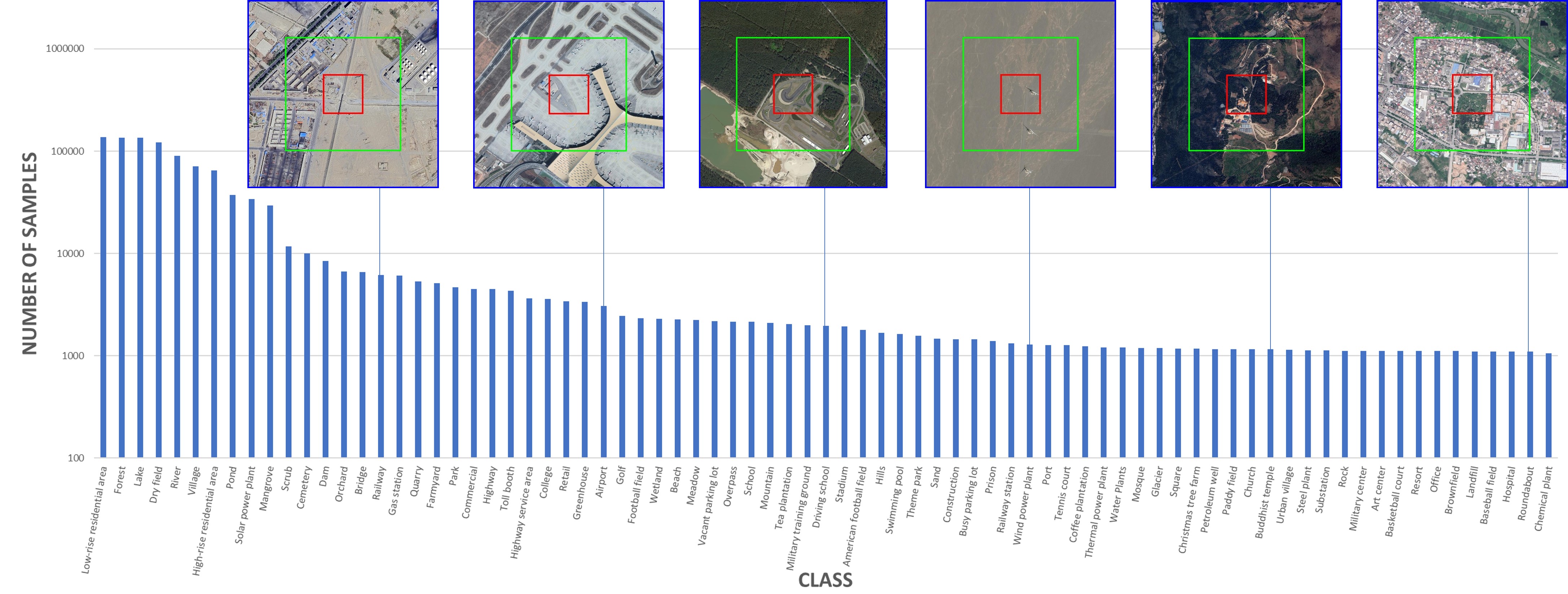}
 \caption{Statistics and visualization of samples from MEET.}
\label{fig:dataset_stat}
 
\end{figure*}

\begin{figure}[t]
\centering
	\includegraphics[scale=.35]{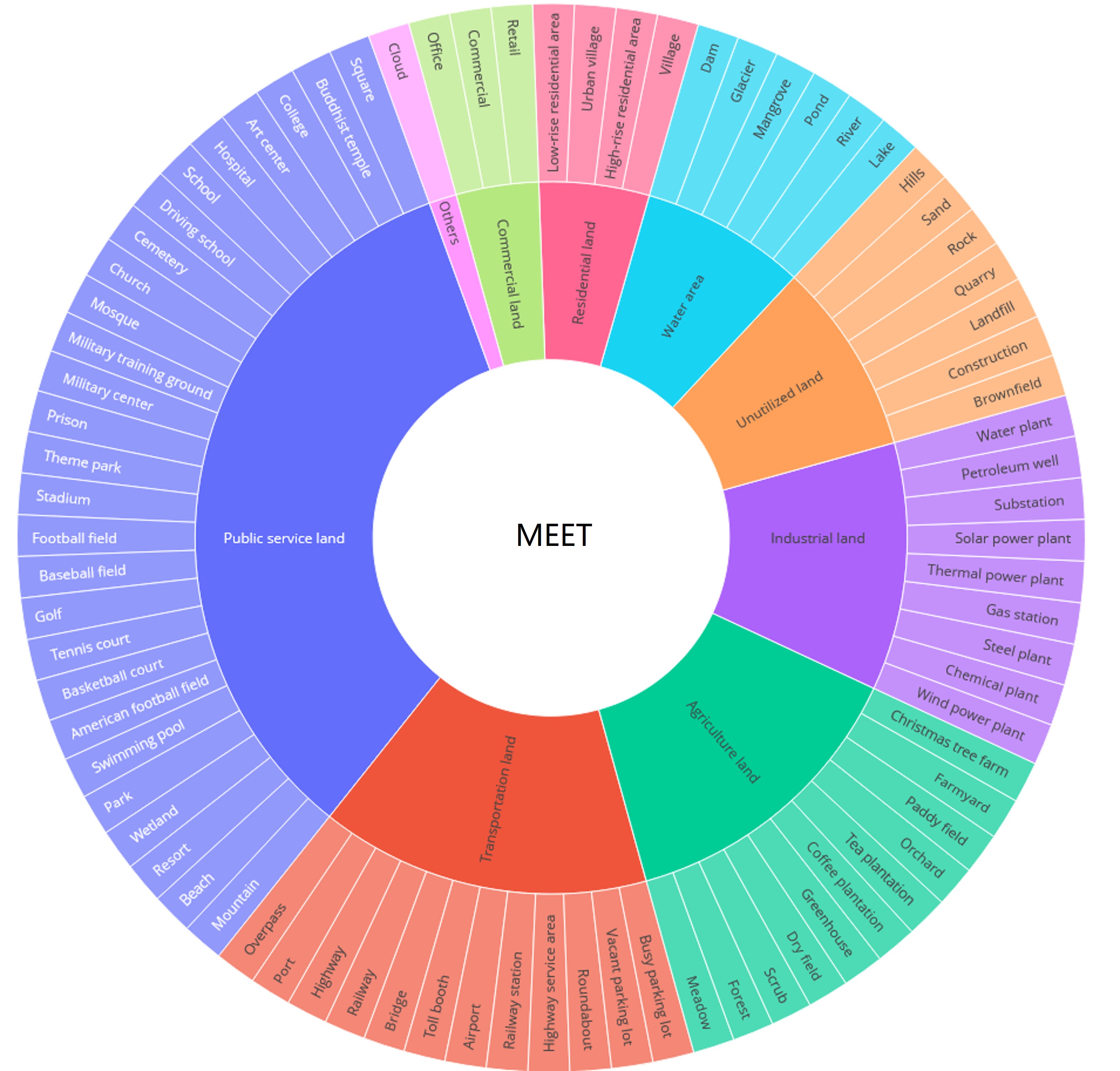}
 \captionsetup{justification=justified, singlelinecheck=false}
\caption{Hierarchical scene category of MEET. All categories are hierarchically organized in a two-level tree: 80 leaf nodes fall into 9 parent nodes, representing 9 underlying scene categories of commercial land, residential land, water area, unutilized land, industrial land, agriculture land, transportation land and public service land.}
\label{fig:class}
\end{figure}
\begin{figure*}
  \centering
  \includegraphics[scale=.5]{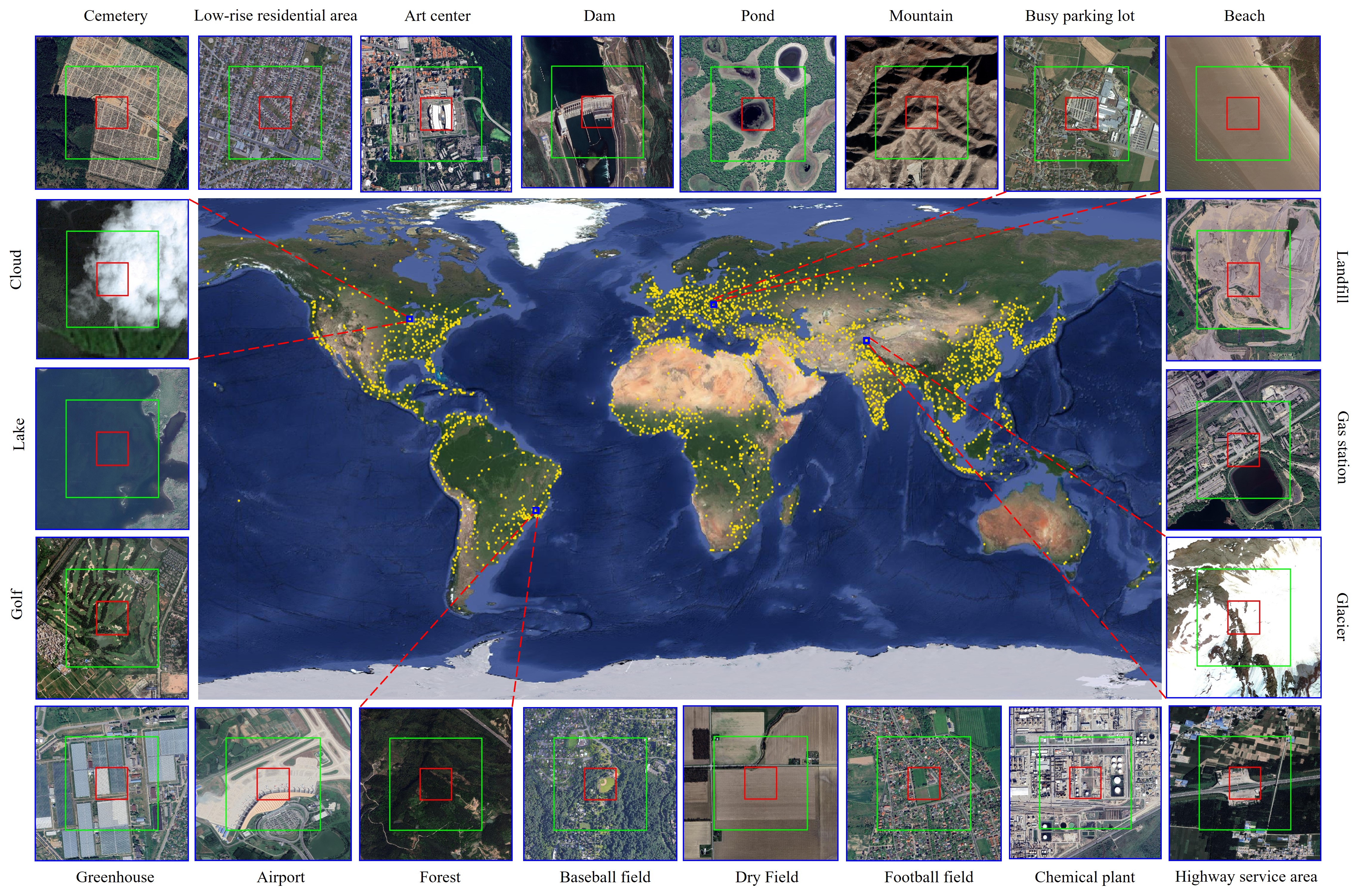}
  \caption{The geographical distribution map of the sampled images from the proposed MEET dataset.}
  \label{fig:distribute}
\end{figure*}

\begin{figure*}
  \centering
  \includegraphics[scale=.48]{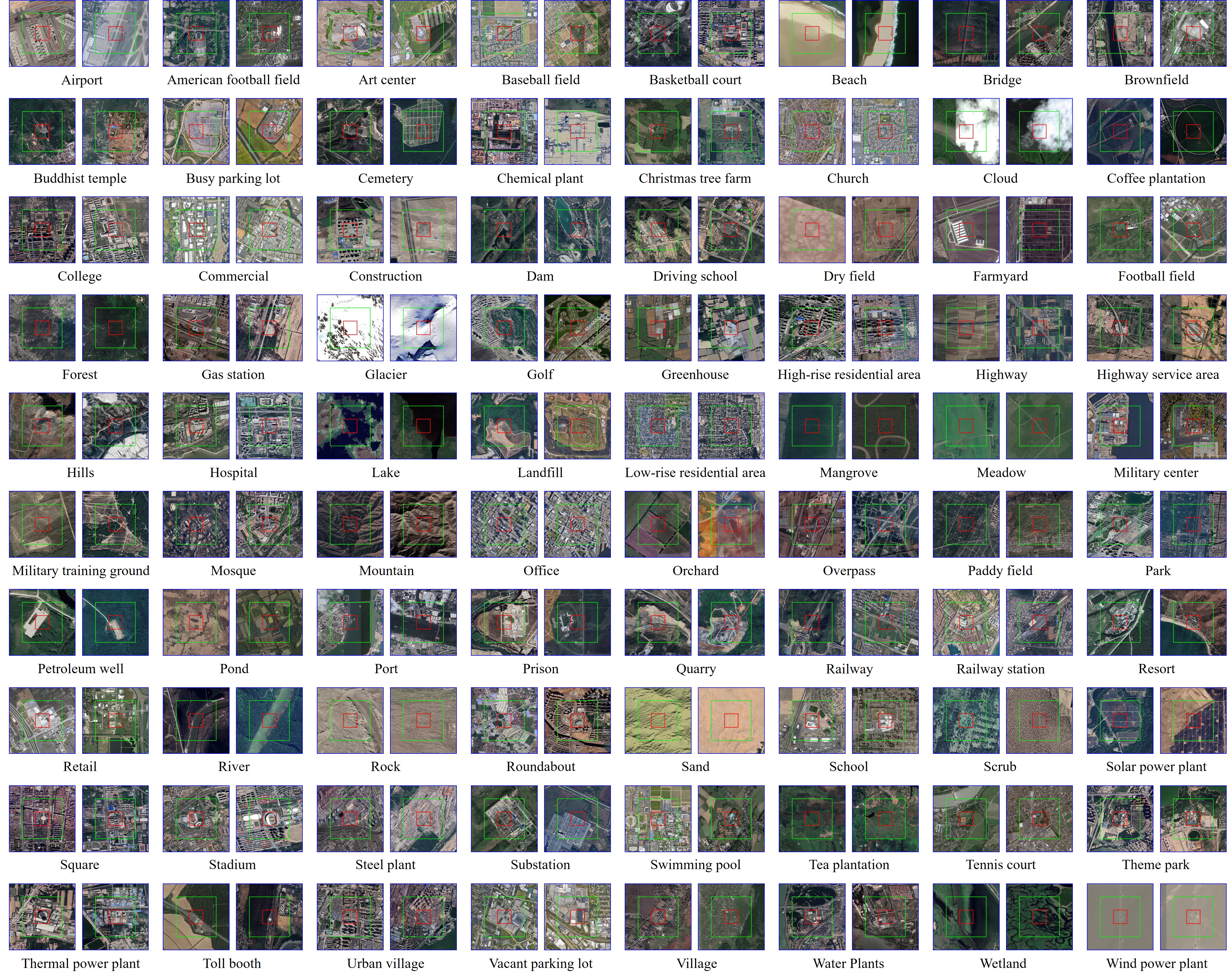}
   \captionsetup{justification=justified, singlelinecheck=false}
  \caption{
  Some examples from the proposed MEET dataset, which employs a scene-in-scene layout. These images exhibit rich variations in appearance, illumination, background, occlusion, and other factors.
  }
  \label{fig:all_case}
\end{figure*}

\section{Related Works}
\label{sec:related_work}

In this section, we provide a concise review of the most pertinent studies in the field, encompassing scene classification datasets, remote sensing scene classification methods and auxiliary image context exploitation methods.

\subsection{Scene Classification Datasets}

As shown in Table~\ref{fig:all_dataset}, the emergence of numerous datasets led to significant advancements in remote sensing scene classification. The earliest dataset in this field was UC Merced~\citep{yang2010bag}. Despite having a relatively small number of samples (only 100 samples per category), it played a crucial role in advancing research on geospatial scene classification tasks. In the subsequent decade, nearly 20 additional remote sensing scene classification datasets were introduced, each contributing to the evolution of the field. The total number of samples in these datasets grew significantly, from a few thousand~\citep{yang2010bag, xia2010structural, zou2015deep, penatti2015deep, zhao2016feature, zhu2016bag, wang2018scene} to tens of thousands~\citep{cheng2017remote, xia2017aid, helber2019eurosat, zhou2018patternnet, li2020rsi}, and even to hundreds of thousands~\citep{basu2015deepsat, qi2020mlrsnet, li2021robust, hua2021multiscene}. This expansion significantly broadened the scope and application scenarios for scene classification tasks. In terms of category diversity, the number of categories also increased over time, from fewer than 10~\citep{zou2015deep, basu2015deepsat, penatti2015deep} to over 40~\citep{cheng2017remote, xiao2017high, sumbul2019bigearthnet, qi2020mlrsnet, long2021creating}. For example, the SR-RSKG dataset~\citep{li2021robust} reached 70 categories, further enhancing the richness of classification tasks. Regarding data sources, most of these datasets were primarily based on Google Earth imagery~\citep{xia2010structural, zou2015deep, zhao2016feature, cheng2017remote, xia2017aid, long2021creating}, while some used freely available medium-resolution satellite imagery, such as Sentinel-2~\citep{helber2019eurosat, sumbul2019bigearthnet}. A smaller subset of datasets utilized data from other sources, including the United States Geological Survey (USGS)~\citep{yang2010bag}, Bing Maps~\citep{li2020rsi, li2020clrs}, and Tianditu~\citep{xiao2017high}.

Despite efforts, existing datasets with zoom-free RSI~\citep{yang2010bag,zhao2016feature,zhu2016bag,xia2017aid,helber2019eurosat,yuan2022wh} were limited in terms of the number of categories. To address this issue, many approaches~\citep{basu2015deepsat,cheng2017remote,xiao2017high,zhou2018patternnet,li2020rsi,qi2020mlrsnet,li2020clrs,long2021creating,li2021robust,hua2021multiscene,li2024star} manually zoomed RSI to construct datasets aimed at improving class separability for FGSC. However, this data-construction technique created a mismatch with practical land-cover applications, which require fixed-resolution imagery. Therefore, developing datasets that incorporate a fine-grained and distinguishable scene category system with zoom-free RSI remained an unaddressed area in previous work.

\subsection{Scene Classification Methods}

To motivate FGSC with zoom-free RSI, in the following paragraphs, we reviewed existing deep learning methods for scene classification with RSI and explored potential technologies to address FGSC challenges using zoom-free RSI. Scene classification has been extensively studied across both natural images and RSI, with various approaches applied to both domains.

For natural images, numerous studies~\citep{yu2024inceptionnext,he2016deep,sun2019deep,tu2022maxvit,ding2022davit,liu2021swin} focused on optimizing the design of general scene classification backbones, which were validated across a wide range of visual downstream tasks. 
Similarly, in the domain of RSI, significant advancements in scene classification were achieved through three primary approaches: (1) training models from scratch~\citep{liu2020arc,bai2022remote,chen2022gcsanet,song2024quantitative}, (2) adapting pre-trained models from ImageNet to RSI~\citep{cheng2017remote,hu2015transferring,li2017integrating}, and (3) fine-tuning pre-trained models specifically for RSI data~\citep{guo2024skysense,chaib2017deep,xu2020two,fang2019robust,xiong2024neural}.
Among the methods involving training from scratch, ARC-Net~\citep{liu2020arc} incorporated residual blocks with asymmetric convolution (RBAC) to reduce computational cost and shrink the model size. Additionally, dilated convolutions and multi-scale pyramid pooling modules were used to expand the receptive field and improve accuracy. Bai et al.~\citep{bai2022remote} proposed a multiscale feature fusion covariance network with octave convolution, which extracted multifrequency and multiscale features from RSIs. Chen et al.~\citep{chen2022gcsanet} introduced GCSANet, which leveraged global context spatial attention (GCSA) and densely connected convolutional networks to capture multiscale global scene features. For methods involving fine-tuning pre-trained models specifically for RSI data, Guo et al.~\citep{guo2024skysense} introduced geo-context prototype learning to learn region-aware prototypes based on RSI’s multi-modal spatiotemporal features. Each of these approaches uniquely enhanced the discriminative ability and robustness of models, driving advancements in the field. For the more challenging task of FGSC, many studies turned to auxiliary data to improve the interpretation of fine-grained categories. Srivastava et al.~\citep{srivastava2020fine} optimized FGSC performance by utilizing visual cues from side-view pictures sourced from Google Street View (GSV). Similarly, Fang et al.~\citep{fang2022spatial} incorporated street view images (SVI) and developed a spatial context-aware land-use classification method to enhance land-use classification accuracy. 
Yao et al.~\citep{yao2022classifying} introduced temporal resolution time-series electricity data to explore the relationship with socioeconomic attributes and constructed a neural network that can fuse time-series electricity data and RSIs to identify urban land-use types.
Arbinger et al.~\citep{arbinger2022exploiting} introduced geographic coordinates or geoinformation data to enable a better understanding of the image content and thus facilitate their classification.
The limited availability and high acquisition cost of additional data sources posed challenges and restricted the broader application of these methods.

\subsection{Auxiliary Image Context Exploitation Methods}

In literature, incorporating auxiliary contextual information was regarded as a natural and effective approach to enhance the interpretability of RSI. In semantic segmentation of RSI, Li et al.~\citep{li2023mfvnet} proposed a deep adaptive fusion network with multi-scale context, specifically designed for RSI semantic segmentation. GLNet~\citep{Chen_2019_CVPR} preserved both global and local information in a highly memory-efficient manner, capturing high-resolution fine structures from zoomed-in local patches and contextual dependencies from the downsampled input. CascadePSP~\citep{cheng2020cascadepsp} used a global step to refine the entire image, providing sufficient image context for a subsequent local step to perform full-resolution, high-quality refinement. In object detection of RSI, HBD-Net~\citep{li2024learning} addressed bridge detection by incorporating multi-scale context within the dynamic image pyramid (DIP) of large-scale images, while employing a shape-sensitive sample re-weighting (SSRW) strategy to balance regression weights for bridges with varying aspect ratios. GLGCNet~\citep{bai2023global} extracted global representations and combined them with local-context-aware features gathered from three saliency-up modules for comprehensive saliency modeling. An edge assignment module was also employed to refine preliminary detections. GeoAgent~\citep{liu2023seeing} enhanced performance by adaptively capturing contextual information based on geographical objects, using a feature indexing module to differentiate locations. However, to the best of our knowledge, no work on scene classification has yet utilized contextual information, let alone for FGSC. Furthermore, these semantic segmentation and object detection methods could not be directly adapted to FGSC with image-level labels. This left the exploitation of contextual information in FGSC as an open and significant research space.

\section{The Proposed MEET Dataset}
\label{sec:glh_dataset}
Our goals for developing a new dataset for FGSC are twofold: (i) to promote a new meaningful yet challenging task: FGSC with zoom-free RSI; (ii) to occupy the niche of FGSC datasets with context image.  This section provides a comprehensive overview of the MEET dataset, focusing on three key aspects: data collection and organization, data annotation, and data analysis.

\subsection{Data Collection and Organization}
We select fixed-resolution samples and supplement them with surrounding imagery as multi-scale context. To ensure data diversity, images are collected globally, covering variation in appearance, illumination and occlusion.
The fine-grained geospatial scene category is determined by the center scene, with auxiliary scenes serving as contextual information. The MEET dataset provides global coverage through the collection of 1.03 million samples spanning Asia, Africa, South America, North America, and Europe, and covering 80 typical scene categories. The images are collected from 2018 to 2022, with each sample containing a center scene with 256 × 256 pixels, along with a surrounding scene with 768 × 768 pixels and a global scene with 1280 × 1280 pixels. The overall distribution and some samples are shown in Fig.~\ref{fig:distribute}.
It is important to note that the spatial resolution of all samples is consistently set to 1.0m.

To comprehensively obtain RSI with diverse and comprehensive scene categories on a global scale, we leverage data from OpenStreetMap (OSM), which is a collaborative project creating a free, editable map of the world. OSM provides semantic annotations by labeling geographical features and land-use within its maps, offering detailed information about roads, buildings, and other points of interest. Subsequently, we preprocess the acquired data by performing coarse filtering or integration of scene categories based on the quality of OSM annotations, and design a series of rules to reduce noise at the image-level labels. Finally, we ensure a global distribution and high richness of samples by defining random spatial windows using geographic coordinates.

We randomly divide the entire dataset into training, validation, and testing sets by category, with a ratio of 6:2:2. More specifically, the training set, validation set, and testing set contain 620,237, 206,755, and 206,755 samples, respectively.

\begin{figure*}[htbp]
    \centering
    \subfigure[Illustration of intra-class variation]{
        \includegraphics[scale=.67]{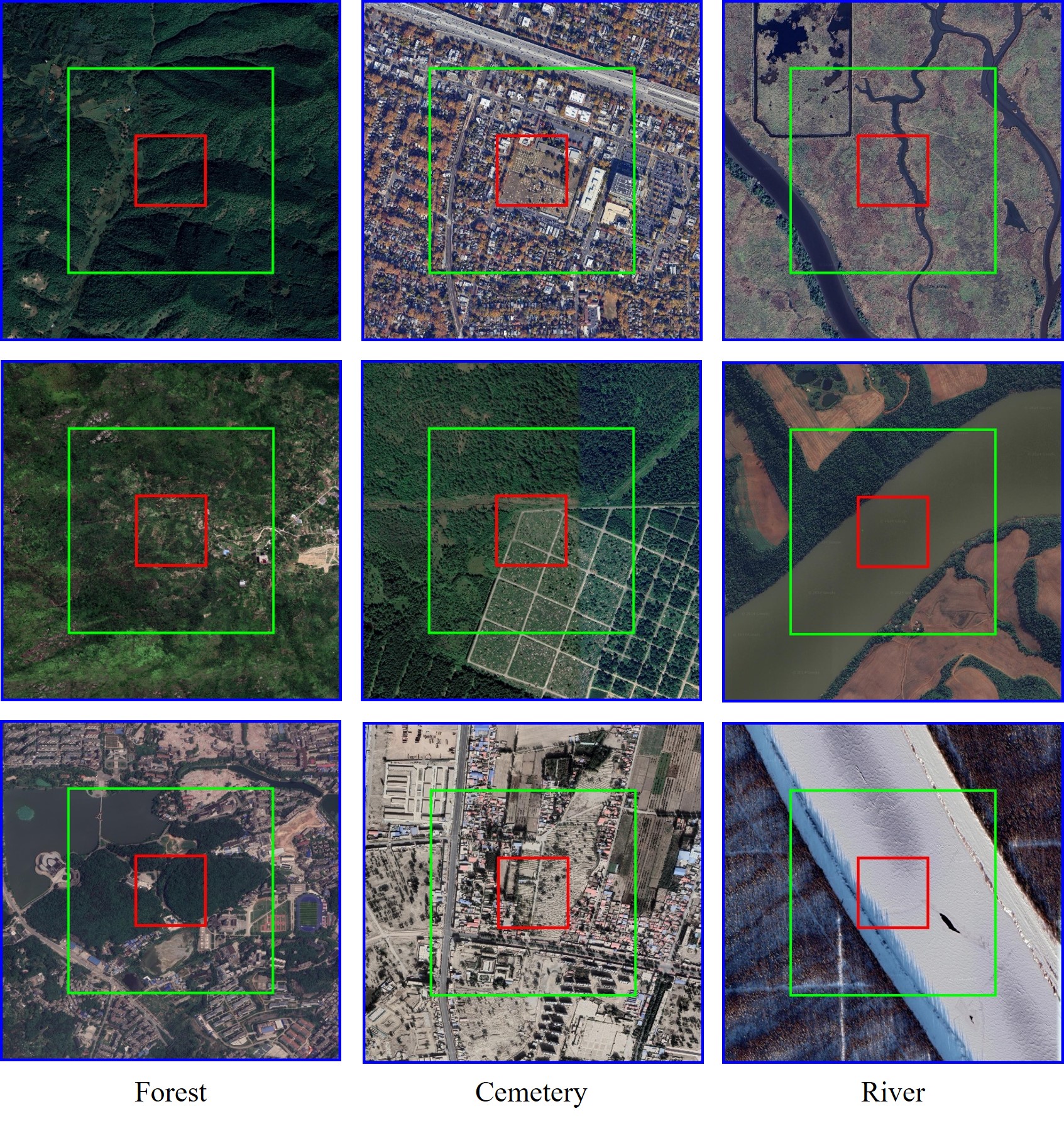}
        \label{fig:intra}
    }
    \subfigure[Illustration of inter-class similarity]{
        \includegraphics[scale=.67]{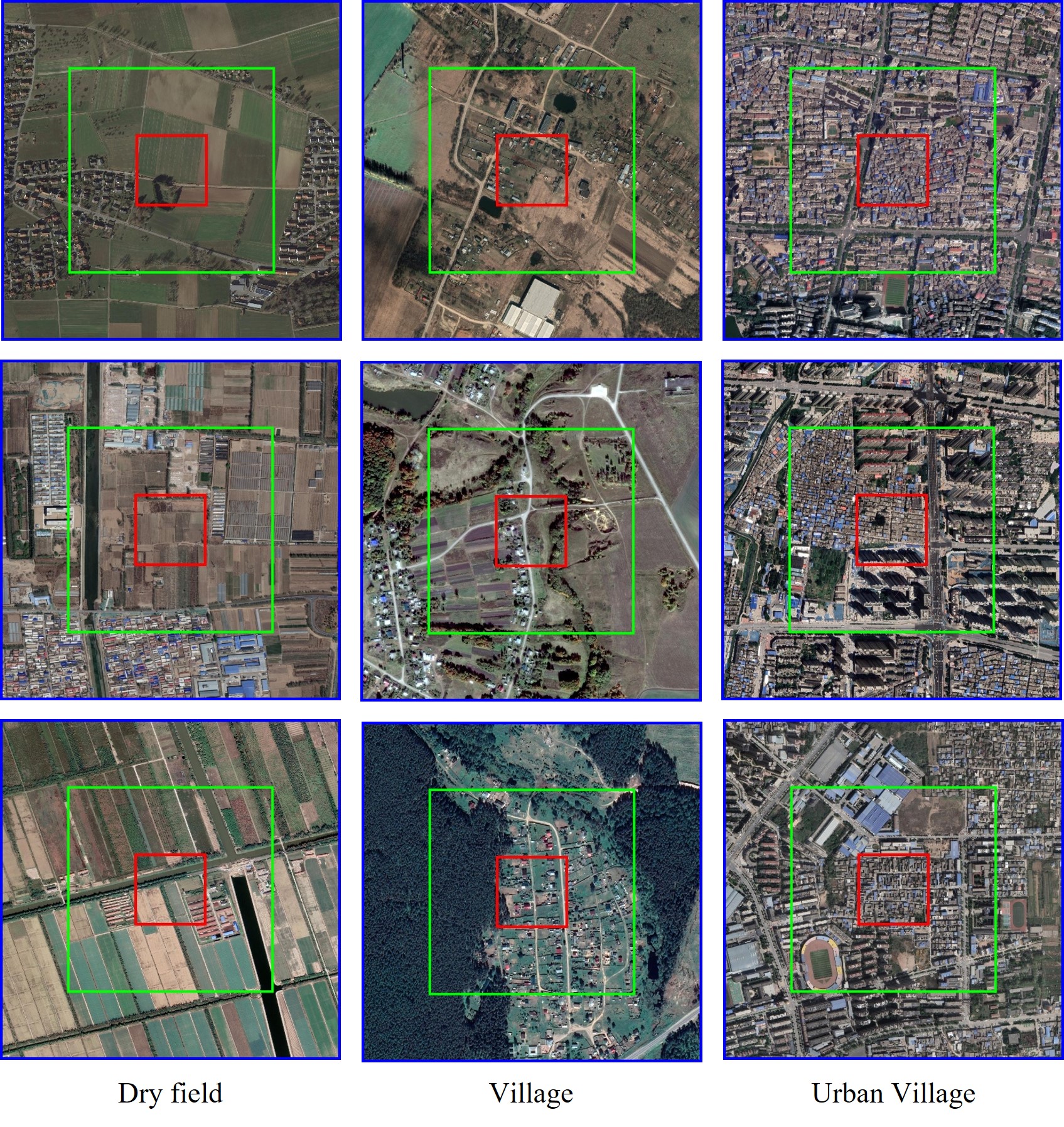}
        \label{fig:inter}
    }
    \caption{Illustration of intra-class variation and inter-class similarity in MEET.}
    \label{fig:intra_inter}
\end{figure*}

\subsection{Data Annotation}
Overall, fine-grained geospatial scene categories are defined by considering the center scene along with its auxiliary scenes.
To ensure precise annotations of MEET dataset, ten trained experts in the field of remote sensing participate in the annotation process with cross-validation.
During the annotation process, each annotator receives the center scene along with the corresponding contextual imagery. For samples whose categories can be determined solely based on the center scene, annotators use the predominant scene classification within the image as the scene label for the current sample. For example, a mosque located within a residential area of a city, due to its smaller footprint and relative rarity, is categorized under \textit{mosque}. This strategy enhances the coverage of smaller target features, thereby enriching the holistic understanding of urban attributes.
As far as samples whose categories require contextual imagery to determine, annotators can also make correct labeling choices by combining the corresponding context. For instance, for texture-poor water body region images, annotators can distinguish whether the sample belongs to the river or lake category by considering the shape of the banks in the surrounding context.

The procedure of labeling MEET encompasses a tripartite framework consisting of three stages: pre-annotation stage, expert feedback and optimization stage, and large-scale detailed annotation stage. In the initial phase of pre-annotation, we form a specialized team comprising 10 members, each possessing extensive expertise in the field of remote sensing interpretation. This team undergoes comprehensive training in fundamental annotation techniques and subsequently conducts annotation tests on a representative subset of the dataset. In the following feedback and optimization stage, experts thoroughly review and evaluate the team’s initial annotations, leading to the formulation of improved annotation standards. Subsequently, guided by these adjustments, the team embarks on the formal large-scale annotation process, accompanied by experts’ random sampling inspections.

\subsection{Dataset Analysis}

The MEET dataset distinguishes itself from existing remote sensing scene classification datasets through several unique attributes: the breadth of its category coverage, the scale of its sample size, the diversity of its samples, and the incorporation of contextual information. Additionally, the dataset maintains a uniform sample resolution and is tailored to support models designed for large-scale scene classification and mapping tasks, further emphasizing its distinctive characteristics.

\textbf{Fine Granularity of Categories}: The MEET dataset comprises 80 fine-grained geospatial scene categories, categorized into 11 major scene types. With the introduction of auxiliary contextual information, it has become possible to annotate more fine-grained categories. These categories comprehensively cover discernible remote sensing scene categories. Therefore, our MEET dataset offers advantages over existing remote sensing scene classification datasets by providing more high-value fine-grained scene categories. Especially in urban mapping and analysis applications, these fine-grained scene categories make a wider range of scene classification applications possible.

\textbf{Large Volume of Samples}: The MEET dataset includes 1,033,801 samples, covering over 3.3 billion square kilometers globally. It surpasses other publicly available datasets in both sample volume and richness of annotation data for remote sensing scene classification.

\textbf{High Intra-class Variability and Inter-class Similarity}:
Intra-class variation is mainly due to differences in appearance. Inter-class similarity arises from similar appearance representations in the center scene, but it manifests differently in auxiliary scenes. As shown in Fig.~\ref{fig:intra}, samples in the river category exhibit high richness in image quality, color variations, seasonal changes, geographic regions, and river widths. Conversely, inter-class similarities underscore the utility of contextual information in enhancing classification accuracy, as illustrated in Fig.~\ref{fig:inter} where incorporating context aids in discerning challenging objects within the current block.

Although the distribution of the MEET dataset exhibits a certain degree of class imbalance, as shown in Fig.~\ref{fig:dataset_stat}, it closely mirrors the frequency distribution of real-world scene categories. This characteristic enhances its value for practical applications. Additionally, it is important to emphasize that the selected samples exhibit significant variation within each category, ensuring that even among the head classes, homogeneous low-value samples are also relatively few.


\section{Proposed Method}
\label{sec:proposed_method}
\begin{figure*}[t]
\centering
\includegraphics[scale=.5]{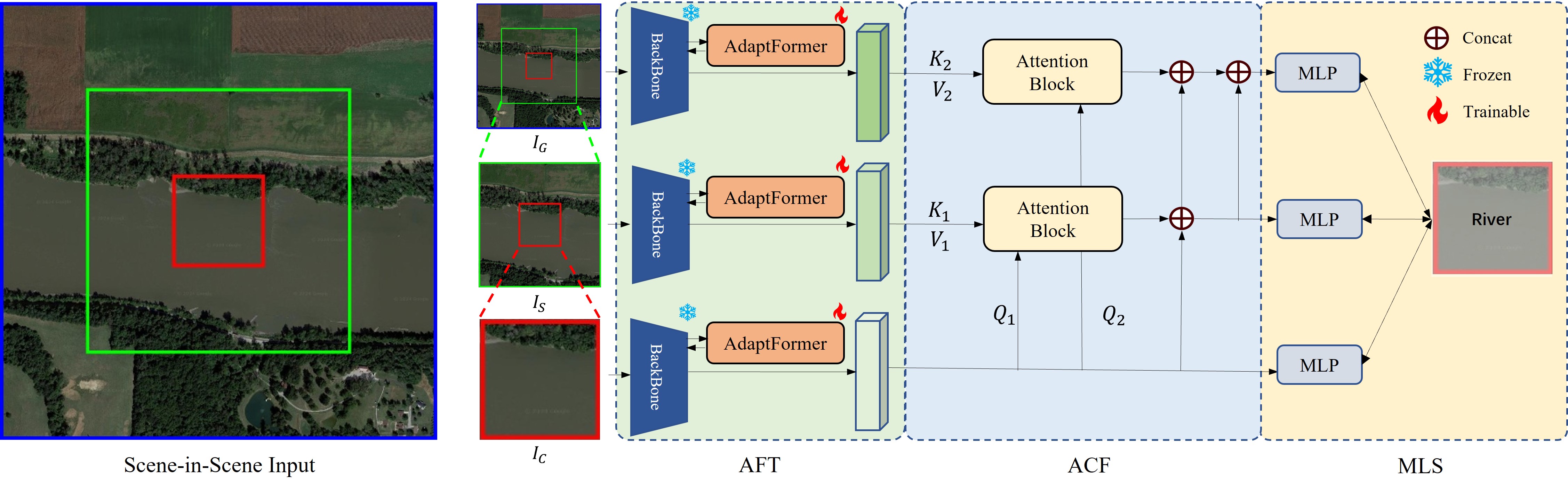}
\captionsetup{justification=justified, singlelinecheck=false}
\caption{Overview of CAT. The structure contains three components (from left to right): AdaptFormer Tuning (AFT), Adaptive Context Fusion (ACF), and Multi-level Supervision Optimization (MLS).}
\label{fig:model}
\end{figure*}

To flexibly and efficiently exploit the scene-in-scene layout in FGSC with zoom-free RSI, this paper introduces CAT, a novel approach specifically tailored for this task. CAT incorporates an adaptive context fusion module to effectively extract multi-scale contextual features from the transformer backbone. To ensure performance without excessively increasing parameters, we utilize parameter-efficient fine-tuning (PEFT) methods to finetune the backbone, instead of training from scratch or parameter synchronization. Additionally, we introduce multi-level supervision through independent classification heads during training. This improves feature learning at each level and mitigates overfitting that can arise from auxiliary scenes. 
More specifically, we utilize the scene-in-scene layout for each sample with large range of context.
However, contemporary deep networks face limitations in directly processing large-size RSI due to GPU memory constraint.
To address this challenge, we resize each image to a uniform size, denoted as \( I_C \), \( I_S \) and \( I_G \) , respectively.
Among them, \( I_C \) corresponds to center scene, while \( I_S \) and \( I_G \)  correspond to the surrounding and global scene, respectively. During both the training and inference stages, the input to CAT remains consistent. The whole architecture of our method is illustrated in Fig.~\ref{fig:model}. This section is dedicated to provide a detailed explanation of the CAT.

\subsection{AdaptFormer Tuning}
For the currently constructed scene-in-scene layout, three branches are needed to perform feature extraction respectively while using pre-trained weights.
Although efficiency has been improved by resizing input images, using three branches for feature extraction still poses difficulties.
Using completely independent backbones for full-parameter training would significantly increase the model's parameter size, which is unacceptable given the current trend towards larger model parameters.
While using shared weights does not introduce additional parameters, it would hinder performance because the three branches have inputs with fixed but different spatial resolutions.
To address this issue, we introduce the AdaptFormer Tuning (AFT) on the transformer backbone for multi-level image feature extraction.
AdaptFormer replaces the MLP modules in the transformer encoder with AdaptMLP. The computation of the AFT module in transformer can be expressed as follows:
\begin{align}
    z_i &= \text{W-MSA}(\text{LN}(z_{i-1})) + z_{i-1}; \\
    \hat{z}_i &= \text{MLP}_{AFT}(\text{LN}(z_i)) + z_i; \\
    \hat{z}_{i+1} &= \text{SW-MSA}(\text{LN}(\hat{z}_i)) + \hat{z}_i; \\
    z_{i+1} &= \text{MLP}_{AFT}(\text{LN}(\hat{z}_{i+1})) + \hat{z}_{i+1}.
\end{align}
where \( z_i \)  denotes the feature output of the i-th module in transformer.
For each branch, the backbone uses shared weights and is initialized with the pre-trained model's weights, while independent AdaptFormer modules are used for training in the three branches. During training, we freeze the weights from the pre-trained model and only update the weights of the AdaptFormer. With this design, the three branches use the AFT method to enable feature extraction from input images of different spatial resolutions. At the same time, the three branches share most of the parameters, ensuring that the total parameter size does not increase significantly, thus maintaining the model's efficiency and practicality.

\begin{table*}[ht]
\centering
\renewcommand{\arraystretch}{1.3}
\caption{PERFORMANCE (\%) COMPARISON OF THE PROPOSED CAT AND OTHER METHODS \\ON THE MEET DATASET.}
\label{tab:method_comparison}
\begin{tabular}{lccccccccc}
\hline

\hline
\multirow{2}{*}{Method} & \multirow{2}{*}{Type} & \multirow{2}{*}{Backbone} & \multirow{2}{*}{Pretrain Dataset} & \multirow{2}{*}{Input} & \multicolumn{5}{c}{Performance Metrics} \\ \cline{6-10}
 &  &  &  &  & OA & BA\textsubscript{many} & BA\textsubscript{med} & BA\textsubscript{few}  & BA       
\\ \hline

\multicolumn{10}{l}{\textit{Methods Generally Proposed for Natural Image Recognition}} \\

\multirow{2}{*}{InceptionNext~\citep{yu2024inceptionnext}} & \multirow{2}{*}
{CNN} & \multirow{2}{*}{InceptionNext-Base} & \multirow{2}{*}{ImageNet-1K} & Center scene 
& 90.37&93.66&67.99&66.71&71.02 \\
 &  &  &  & Global scene & 92.98&96.72&73.00& 62.88&72.34 \\

\multirow{2}{*}{Resnet~\citep{he2016deep}} & \multirow{2}{*}{CNN} & \multirow{2}
{*}{Resnet101} & \multirow{2}{*}{ImageNet-22K} & Center scene & 
81.58&91.21& 52.71&47.44&55.96 \\
 &  &  &  & Global scene & 82.80&95.46&51.53& 39.57&52.94 \\

\multirow{2}{*}{HRNet~\citep{sun2019deep}} & \multirow{2}{*}{CNN} & \multirow{2}
{*}{HRNet-w64} & \multirow{2}{*}{ImageNet-22K} & Center scene & 
91.66&93.93&71.49& 73.25&75.26 \\
 &  &  &  & Global scene & 94.27&96.52&78.80&67.25&76.76 \\

\multirow{2}{*}{MaxViT~\citep{tu2022maxvit}} & \multirow{2}{*}{Transformer} & 
\multirow{2}{*}{Maxvit-Large} & \multirow{2}{*}{ImageNet-22K} & Center scene & 
91.22 & 94.94 & 69.43 & 69.79 & 73.08 \\
 &  &  &  & Global scene & 93.57 & 97.34 & 74.72 & 65.90 & 74.41 \\

\multirow{2}{*}{DAVit~\citep{ding2022davit}} & \multirow{2}{*}{Transformer} & 
\multirow{2}{*}{Davit-Base} & \multirow{2}{*}{ImageNet-22K} & Center scene & 
90.85&94.64&69.02&66.54&71.58 \\
 &  &  &  & Global scene & 94.26&97.51& 76.65& 68.91&76.52 \\

\multirow{2}{*}{Swin~\citep{liu2021swin}} & \multirow{2}{*}{Transformer} & \multirow{2}{*}{Swin-Large} & \multirow{2}{*}{ImageNet-22K} & Center scene & 
92.23 & 94.89 & 73.95 & 71.24 & 75.78\\
 &  &  &  & Global scene & 95.58&97.82&\underline{83.04}&73.82&81.50 \\ \hline

\multicolumn{10}{l}{\textit{Methods Specifically Proposed for RSI Scene Classification}} \\

\multirow{2}{*}{ARCNet~\citep{liu2020arc}} & \multirow{2}{*}{CNN} & \multirow{2}{*}{Resnet101} & \multirow{2}{*}{--} & Center scene & 88.55&93.27&63.04&59.94&65.99 \\
 &  &  &  & Global scene & 89.78&96.18&65.59&52.84&64.85 \\

 
\multirow{2}{*}{MF2CNet~\citep{bai2022remote}} & \multirow{2}{*}{CNN} & 
\multirow{2}{*}{Resnet50} & \multirow{2}{*}{--} & Center scene & 
67.52&85.75&31.90&25.18&36.70 \\
 &  &  &  & Global scene & 88.43&88.43&34.93&20.39&36.65 \\

\multirow{2}{*}{GCSANet~\citep{chen2022gcsanet}} & \multirow{2}{*}{CNN} & 
\multirow{2}{*}{Densenet121} & \multirow{2}{*}{--} & Center scene & 
85.44&92.39&59.04&54.10&61.71 \\
 &  &  &  & Global scene & 89.04&95.72&63.55&50.39&62.88 \\


\multirow{2}{*}{DOFA~\citep{xiong2024neural}} & \multirow{2}{*}{Transformer} & 
\multirow{2}{*}{Vit-Large} & \multirow{2}{*}{DOFA} & Center scene & 
94.88&96.88&80.67&71.41&79.31 \\
 &  &  &  & Global scene & 93.31&94.37&77.52&78.97&80.40 \\

\multirow{2}{*}{SkySense~\citep{guo2024skysense}} & \multirow{2}{*}{Transformer} & \multirow{2}{*}{Swin-Huge} & \multirow{2}{*}{SkySense-21.5M} & Center scene & 94.52 
&97.55&79.57&73.76&79.79 \\
 &  &  &  & Global scene & 94.93&\underline{98.41}&81.43&67.72&78.45

\\

\hline

Our CAT        
& Transformer 
& Swin-Large      
& ImageNet-22K    
& Scene-in-scene
&\underline{95.87} & 97.04&82.14	& \underline{80.05}	& \underline{83.38}	\\
Our CAT
& Transformer 
& Swin-Huge      
& SkySense-21.5M     
& Scene-in-scene
&\textbf{97.74}&\textbf{99.00}&\textbf{90.80}&\textbf{84.19}&\textbf{89.37} \\ 
\hline

\end{tabular}
\end{table*}

\subsection{Adaptive Context Fusion}

The model uses a backbone combined with AFT to extract features from 
\( I_C \), \( I_S \) and \( I_G \) , denoted as \( F_C \), \( F_S \) and \( F_G \)  , respectively. Among them, feature \( F_C \) contains rich semantic information most relevant to the labels, while \( F_S \) and \( F_G \) serve as contextual features supplementing \( F_G \). 

These contextual features often contain redundant information and are not entirely correlated with the labels. As mentioned in~\citep{zhang2024learn}, excessive redundant contextual information can impair classification results for certain samples. Therefore, inspired by the design of multi-head self-attention modules, we propose the Adaptive Context Fusion (ACF) module to adaptively integrate features from the center scene with either the surrounding scene or the global scene.
For the features extracted from each level of contextual images, the most relevant and valuable features associated with the center scene are further extracted, reducing the redundant information brought by large-scale geographic areas.

Specifically, we employ two multi-head attention modules for adaptive contextual image feature fusion on surrounding scene and global scene. The query feature \( F_C \) retrieves features from the current block, while the keys and values, derived from \( F_S \) or \( F_G \) , are obtained from the corresponding contextual blocks. This process facilitates adaptive feature extraction from the context based on the visual feature of the current block, thereby enhancing focus on the most relevant features.
The ACF Module is implemented using the MultiHeadAttention module. The \( F^S_{ACF} \) and \( F^G_{ACF} \)
are defined as follows:
\begin{align}
\mathbf{F}^{\text{S}}_{\text{ACF}} &= \text{ACF}(\mathbf{F}^{\text{C}}, \mathbf{F}^{\text{S}})
\\
\mathbf{F}^{\text{G}}_{\text{ACF}} &= \text{ACF}(\mathbf{F}^{\text{C}}, \mathbf{F}^{\text{S}},\mathbf{F}^{\text{G}})
\end{align}
where \(F^{S}_{ACF}\) and \(F^{G}_{ACF}\) are the high-value visual features adaptively extracted from the surrounding scene and global scene, respectively, based on the center scene.
For the median branch, features \( F^C_{ACF} \) and \( F^S_{ACF} \)  are concatenated to obtain \( F^S_{Fused} \). For the global branch, features \( F_C \), \( F^S_{ACF} \), \( F^G_{ACF} \)  are concatenated 
to obtain \( F^G_{ACF} \) . These factors can be expressed as:
\begin{align}
\mathbf{F}^{\text{S}}_{\text{fused}} &= \text{Concat}(\mathbf{F}^{\text{C}}_{\text{ACF}}, \mathbf{F}^{\text{S}}_{\text{ACF}})
\\
\mathbf{F}^{\text{G}}_{\text{fused}} &= \text{Concat}(\mathbf{F}^{\text{C}}_{\text{ACF}}, \mathbf{F}^{\text{S}}_{\text{ACF}}, \mathbf{F}^{\text{G}}_{\text{ACF}})
\end{align}

The ACF Module outputs two contextual fusion features, 
\( F^S_{Fused} \) and  \( F^G_{ACF} \) , along with the center scene feature \( F_C \). Features at each level contain high-value visual information relevant to the center scene, with \( F_C \) as the primary feature for that scale.

\begin{figure*}[t]
\centering
\includegraphics[scale=.85]{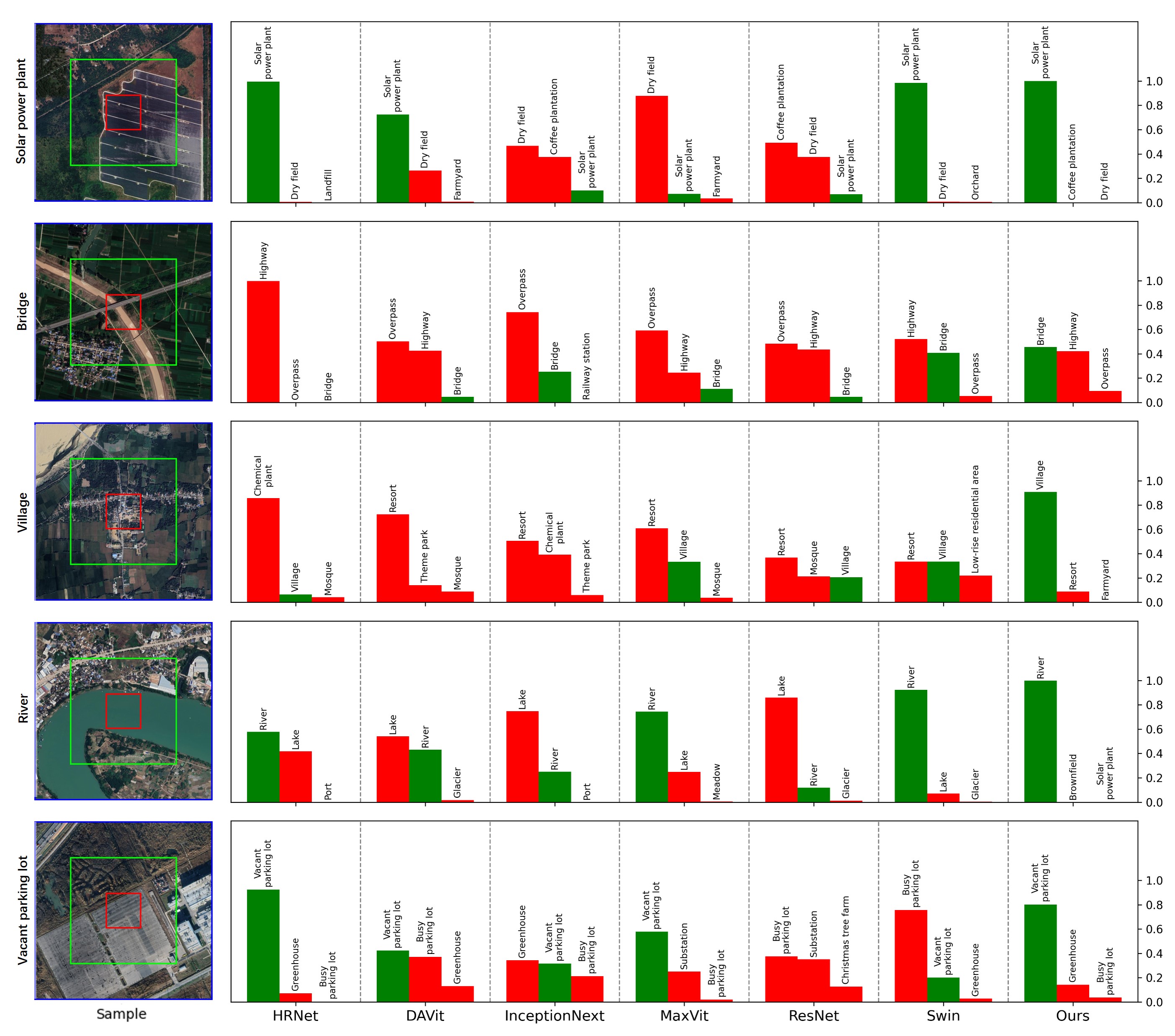}
\captionsetup{justification=justified, singlelinecheck=false}
\caption{Several samples on MEET. Top3 predictions are presented from various comparative methods, as well as those provided by our CAT. Correct prediction categories are displayed in \textcolor{green}{green}, and incorrect prediction categories are displayed in \textcolor{red}{red}.}
\label{fig:case_all}
\end{figure*}

\begin{figure*}[t]
\centering
\subfigure[Input-level fusion]{
		\includegraphics[scale=0.41]{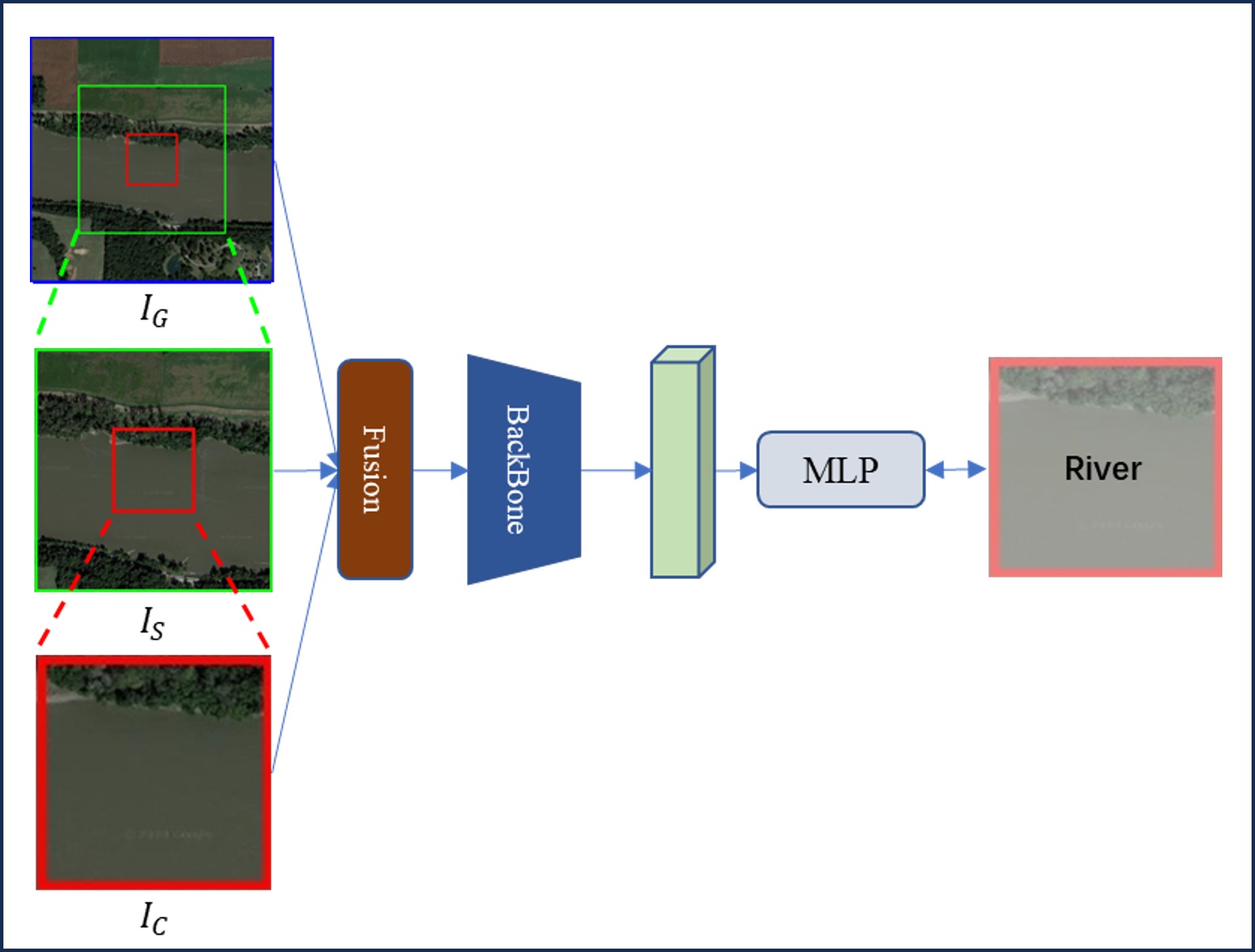}}
\subfigure[Feature-level fusion]{
		\includegraphics[scale=0.41]{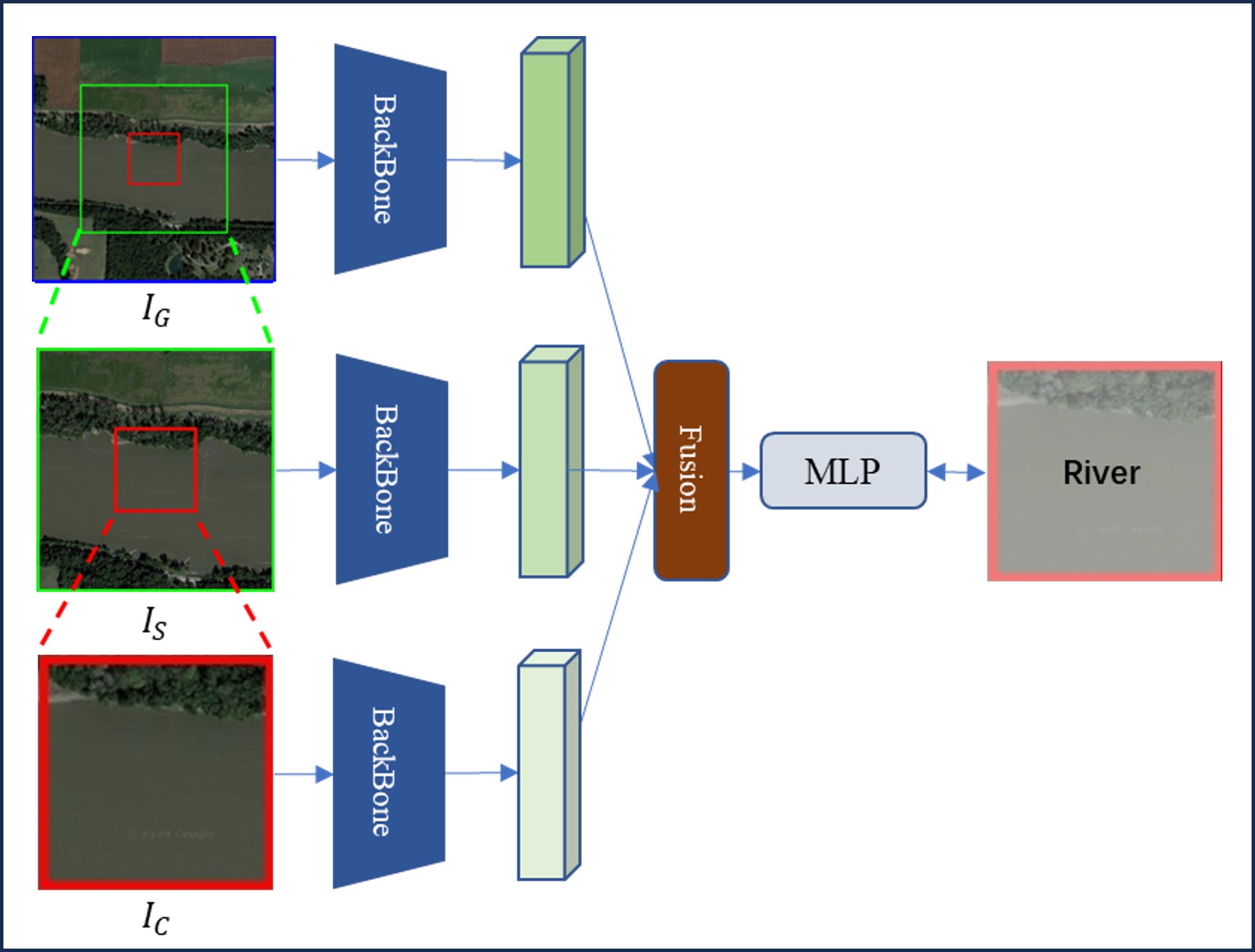}}
\subfigure[Decision-level fusion]{
		\includegraphics[scale=0.41]{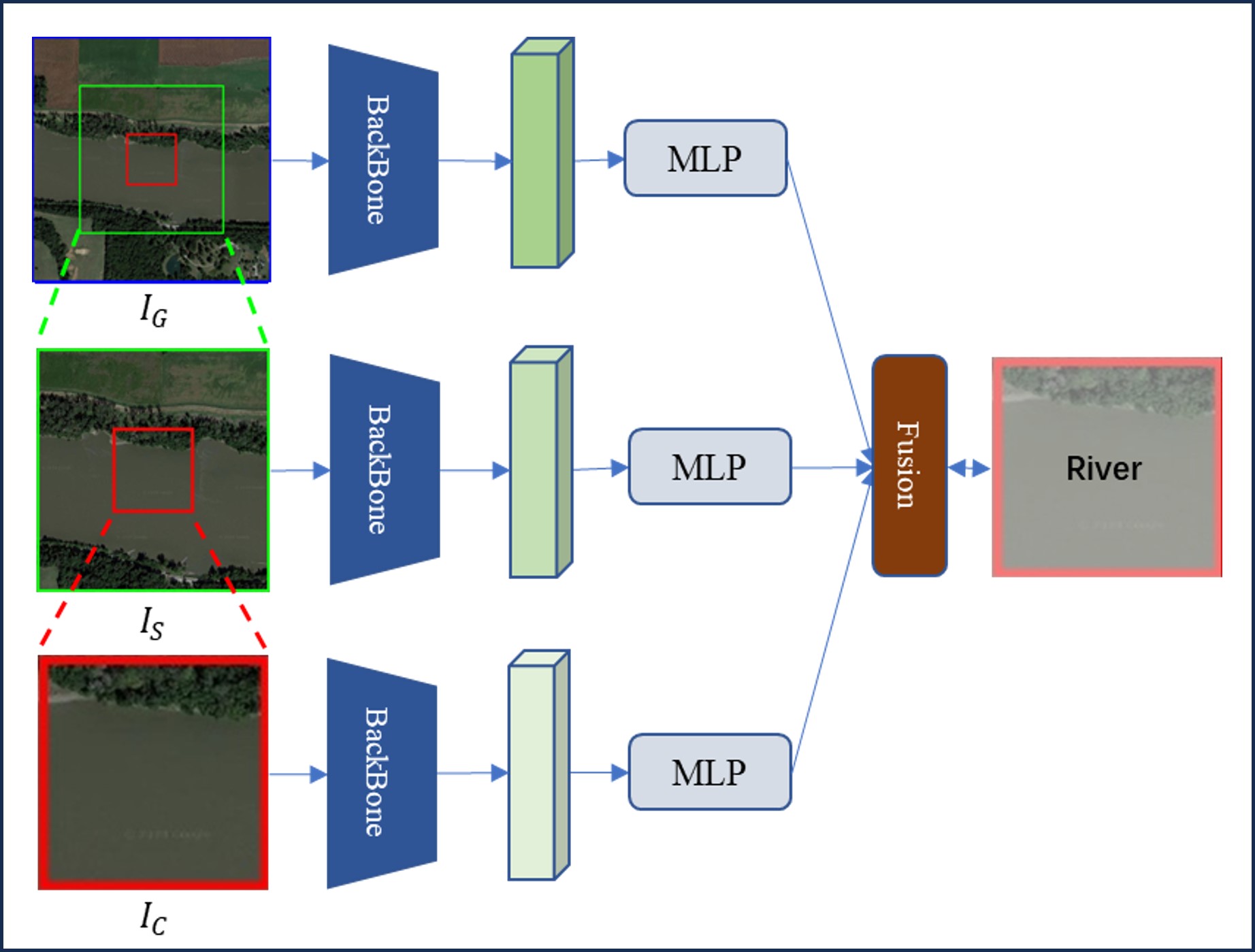}}
\caption{Illustration of three baseline fusion strategies.}
\label{fig:simple}
\end{figure*}

\subsection{Optimization with Multi-level Supervision}

Intuitively, directly utilizing the visual features richest in 
\( F^G_{ACF} \) might achieve the highest classification performance. However, introducing auxiliary scenes may overlook discriminative features of the center scene itself and lead to model overfitting, thus undermining the performance.  
Therefore, utilizing only the branch features rich in contextual information for supervised learning is expected to be insufficient and may also reduce the model's generalization capability.
To address this, we propose a multi-level supervision (MLS) strategy. Specifically, the features extracted from the three branches are subsequently fed into three classification heads for prediction \( P_{C} \), \( P_{S} \) and \( P_{G} \). These factors can be expressed as:
\begin{align}
\mathbf{P}_{\text{C }} &= \text{HEAD}_{\text{C}}(F_C) \\
\mathbf{P}_{\text{S}} &= \text{HEAD}_{\text{S}}(\mathbf{F}^{\text{S}}_{\text{fused}}) \\
\mathbf{P}_{\text{G}} &= \text{HEAD}_{\text{G}}(\mathbf{F}^{\text{G}}_{\text{fused}})
\end{align}

MLS strategy uses ground truth to constrain predictions from all branches to calculate the loss. This ensures that the model extracts effective features even at smaller field-of-view, thereby reducing the risk of associating category semantics with erroneous visual features from the context, thus preventing overfitting.
The total loss of FGSC is defined as follows:
\begin{equation}
\text{Loss}_{all} = \text{Loss}_C+ \text{Loss}_S + \text{Loss}_G
\end{equation}
During inference, we use \( P_{G} \) as the model's predictor, which possesses the complete auxiliary scene information and gets sufficient generalization by MLS.


\section{EXPERIMENTAL RESULTS AND ANALYSIS}
\label{sec:experiment}

In this subsection, we first introduce the evaluation metrics, and then describe our implementation details and mainstream methods for FGSC. Finally, extensive evaluation of our proposed CAT are performed on the MEET dataset.

\subsection{Evaluation Metrics}

Considering the natural long-tail distribution of different scene categories in real-world scenarios, this study uses Overall Accuracy (OA) and Balance Accuracy (BA) as the primary evaluation metrics.
The overall accuracy (OA) is defined as the number of correctly classified images divided by the total number of images in the dataset. The score of OA reflects the overall performance of classification models instead of per class as follows:
\begin{equation}
\text{OA} = \text{N}_c / \text{N}_t
\end{equation}
where \(\text{N}_c\) represents the number of correctly classified images, and \(\text{N}_t\) represents the total number of images in the dataset.
The balance accuracy (BA) is defined as the average OA across all classes in the dataset. The BA score reflects the average performance of the classification model across each class as follows:
\begin{equation}
\text{BA} = \frac{1}{C} \sum_{i=1}^{C} \text{OA}_i
\end{equation}
where \(\text{OA}_i\) represents the OA of the i-th class.
To further understand the performance on the dataset, we categorize the MEET dataset based on sample quantities. Specifically, we define a set of sample ranges as {(0,1500], (1500,10000], (10000,150000]}. Categories are classified based on their sample counts into Many, Medium (Med), and Few. The corresponding BAs are denoted as BA\textsubscript{many}, BA\textsubscript{med}, and BA\textsubscript{few}.

\begin{figure*}[htbp]
    \makeatletter
    \renewcommand{\@thesubfigure}{\hskip\subfiglabelskip}
    \makeatother
    \centering
    \subfigure{
    \includegraphics[width=1\linewidth]{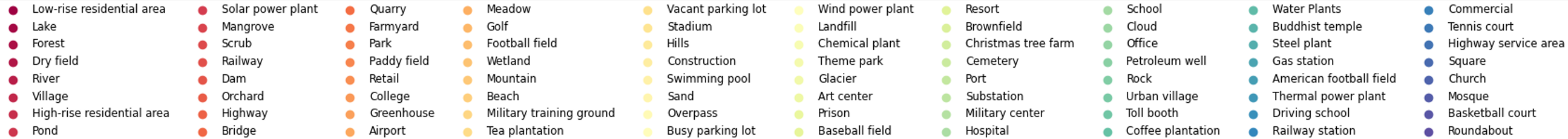}
    }
    \subfigure[(a) Swin]{
        \includegraphics[width=0.48\linewidth]{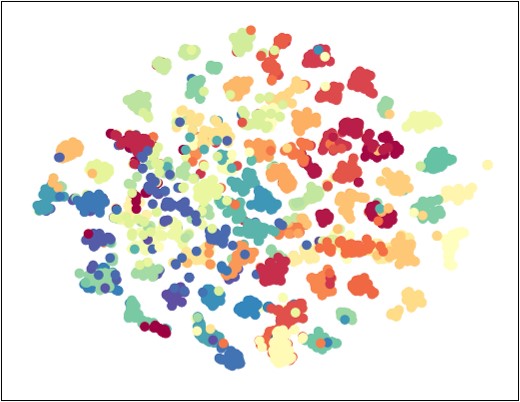}
        \label{fig:top_left_image}
    }
    \subfigure[(b) Swin + ACF]{
        \includegraphics[width=0.48\linewidth]{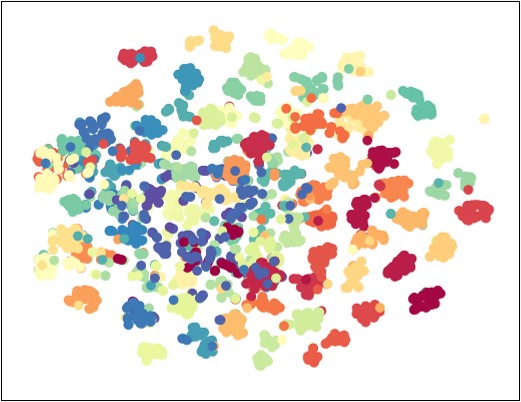}
        \label{fig:top_right_image}
    }


    \subfigure[(c) Swin + ACF + MLS]{
        \includegraphics[width=0.48\linewidth]{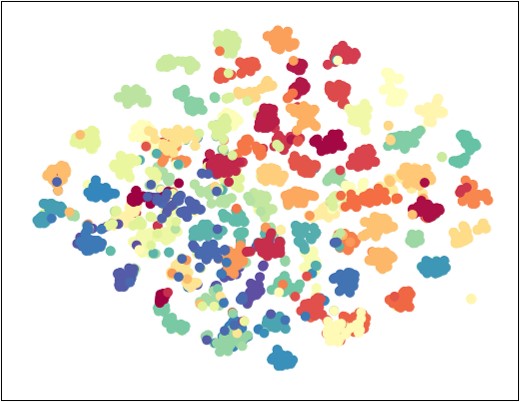}
        \label{fig:bottom_left_image}
    }
    \subfigure[(d) Swin + ACF + MLS + AFT]{
        \includegraphics[width=0.48\linewidth]{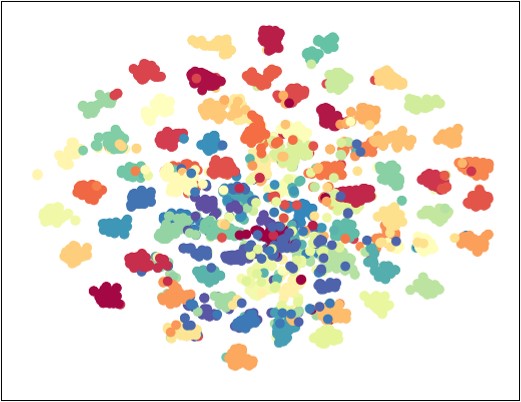}
        \label{fig:bottom_right_image}
    }

    \caption{Visualization of the features for different ablation study settings of our CAT.}
    \label{fig:tsne}
\end{figure*}

\subsection{Implementation Details}
Generally, most algorithms used in our experiments are sourced from the open-source PyTorch-based library TIMM. This library integrates various state-of-the-art computer vision models, along with their respective backbones, feature extractors, and classification heads. They are capable of reproducing the original accuracy of their respective algorithms within a unified framework, ensuring fairness. Additionally, for other models, we use official open-source code as much as possible to ensure experimental rigor.
The experiments are conducted on a server with 1 NVIDIA GeForce RTX 3090 GPU and 24GB of memory. 
To ensure a fair comparison, we apply the most consistent pre-trained model parameters across all methods. 
We use the Adam optimizer with a learning rate of 0.00005. For some remote sensing models, which are relatively smaller, a learning rate of 0.0005 is used to avoid significantly reducing training efficiency, except for SkySense. In all experiments, the batch size is set to 16, except for those using Swin-Huge as the backbone, where it is set to 8.

\subsection{Mainstream Methods}

To establish a benchmark for FGSC with zoom-free RSI, we re-implement scene classification methods. In the remote sensing field, we select several representative works: ARCNet~\citep{liu2020arc}, MF2CNet~\citep{bai2022remote}, GCSANet~\citep{chen2022gcsanet}, DOFA~\citep{xiong2024neural} and SkySense~\citep{guo2024skysense}. Given the rapid progress in exploring backbone models in the general computer vision field, we also incorporate many widely validated methods as strong comparison benchmarks, including ResNet~\citep{he2016deep}, HRNet~\citep{sun2019deep}, Inception-Next~\citep{yu2024inceptionnext}, MaxViT~\citep{tu2022maxvit}, DAVit~\citep{ding2022davit}, and Swin Transformer (Swin)~\citep{liu2021swin}. To ensure the performance of baseline methods, we train all baseline methods with full parameters. While considering the practical usability of our CAT, we use AFT for parameter-efficient fine-tuning, which means performance of CAT could potentially be further improved with full-parameter training.

Considering the substantial benefits of pre-trained model weights for downstream tasks, we use pre-trained model weights on ImageNet-22K for initialization wherever possible, and thus use center scenes or global scenes as model input to meet the three-channel input requirement. For our CAT, since it is specifically designed for scene-in-scene layout, both the center scene and auxiliary scenes are used as inputs.

\begin{table}[t]
\centering
\renewcommand{\arraystretch}{1.3}
\caption{\\PERFORMANCE (\%) COMPARISON OF THE PROPOSED CAT AND OTHER FUSION STRATEGIES}
\label{tab:fusion}
\begin{tabular}{cccccccc}
\hline           
& {STRATEGY} 
& {OA} 
& BA\textsubscript{many}
& BA\textsubscript{med}     
& BA\textsubscript{few}       
& {BA}    
\\ \hline
& Input-level Fusion            
&81.41 & 91.19	&49.31	&39.44	&51.24	
\\
& Feature-level Fusion    
&\underline{86.26}& \underline{95.37}	&\underline{57.99}	&\underline{46.74}	&\underline{58.77}
\\
& Decision-level Fusion        
&85.78& 95.24	&56.95	&43.46	&56.99

\\ \hline
& Our CAT          
&\textbf{95.87} & \textbf{97.04}&\textbf{82.14}	& \textbf{80.05}	& \textbf{83.38}	
\\ \hline
\end{tabular}
\end{table}

\begin{figure*}
\centering
 \includegraphics[scale=1]{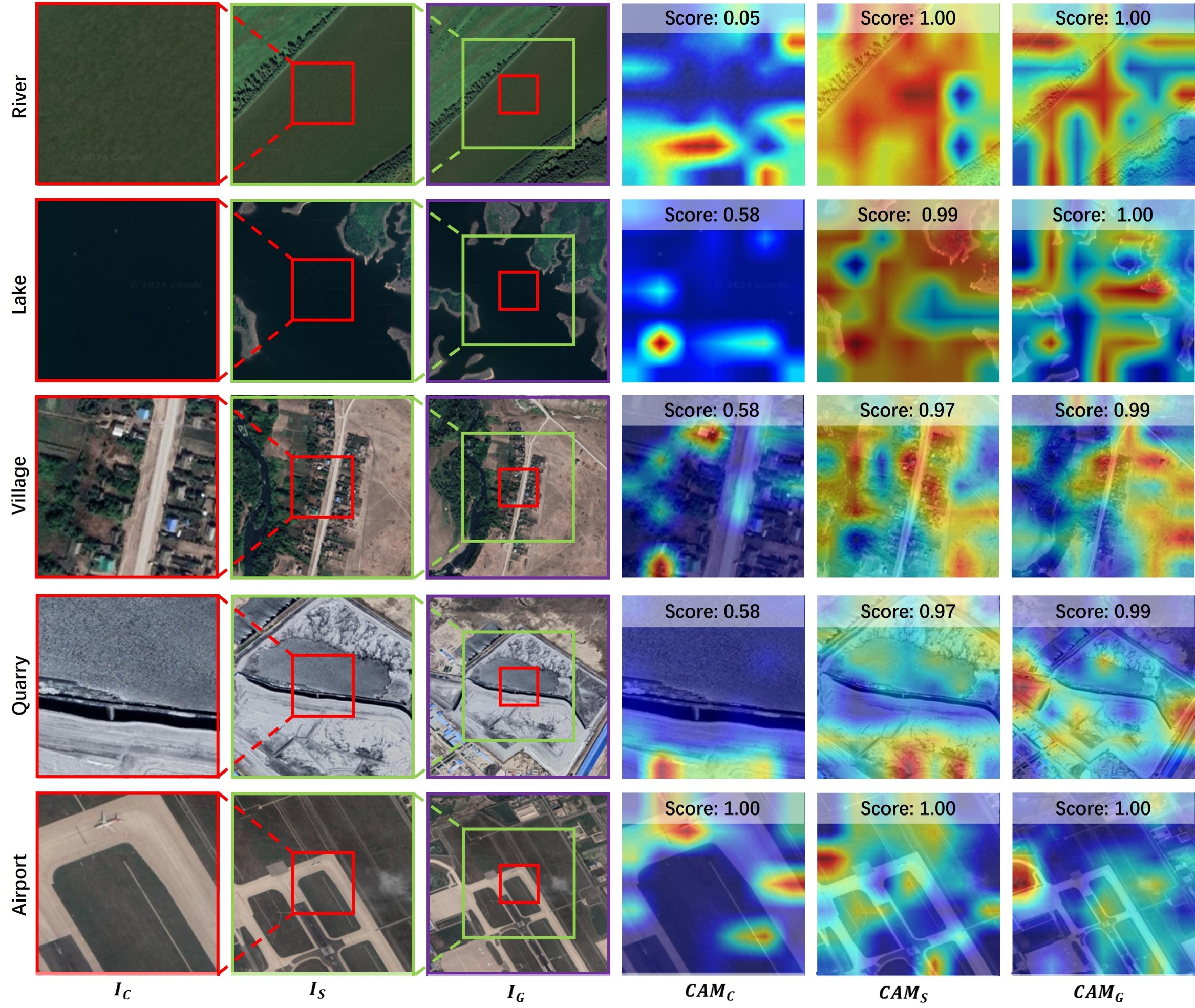}
 \captionsetup{justification=justified, singlelinecheck=false}
\caption{Visualization of the samples using CAM with our CAT. The scores refers to the prediction probability on the ground truth category after applying different amounts of contextual features. Changes in the score reflect the gain in performance due to the accumulation of multi-level context.}
\label{fig:cam}
\end{figure*}

\begin{figure*}[htbp]
    \centering
    \subfigure[Accuracy illustration of Swin]{
        \includegraphics[scale=.24]{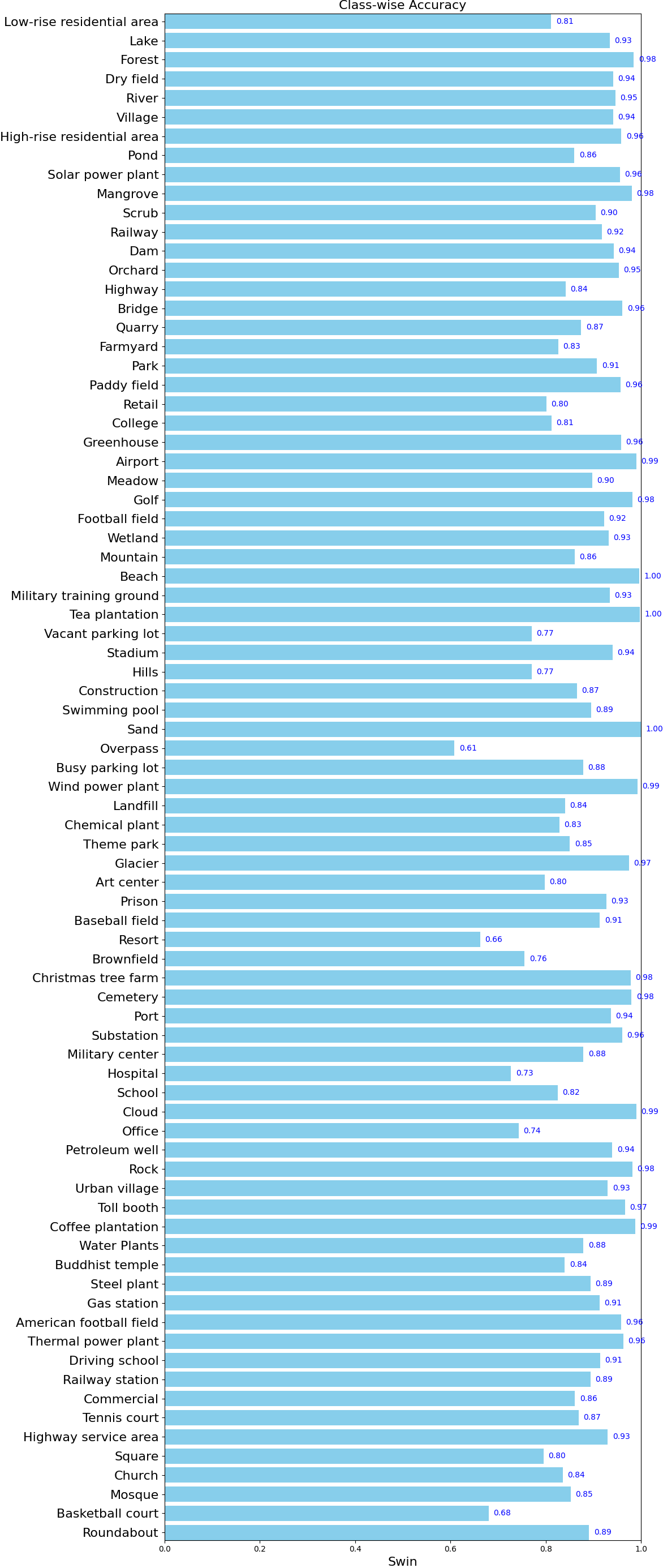}
        \label{fig:class_acc1}
    }
    \subfigure[Accuracy illustration of Swin + ACF + MLS + AFT]{
        \includegraphics[scale=.24]{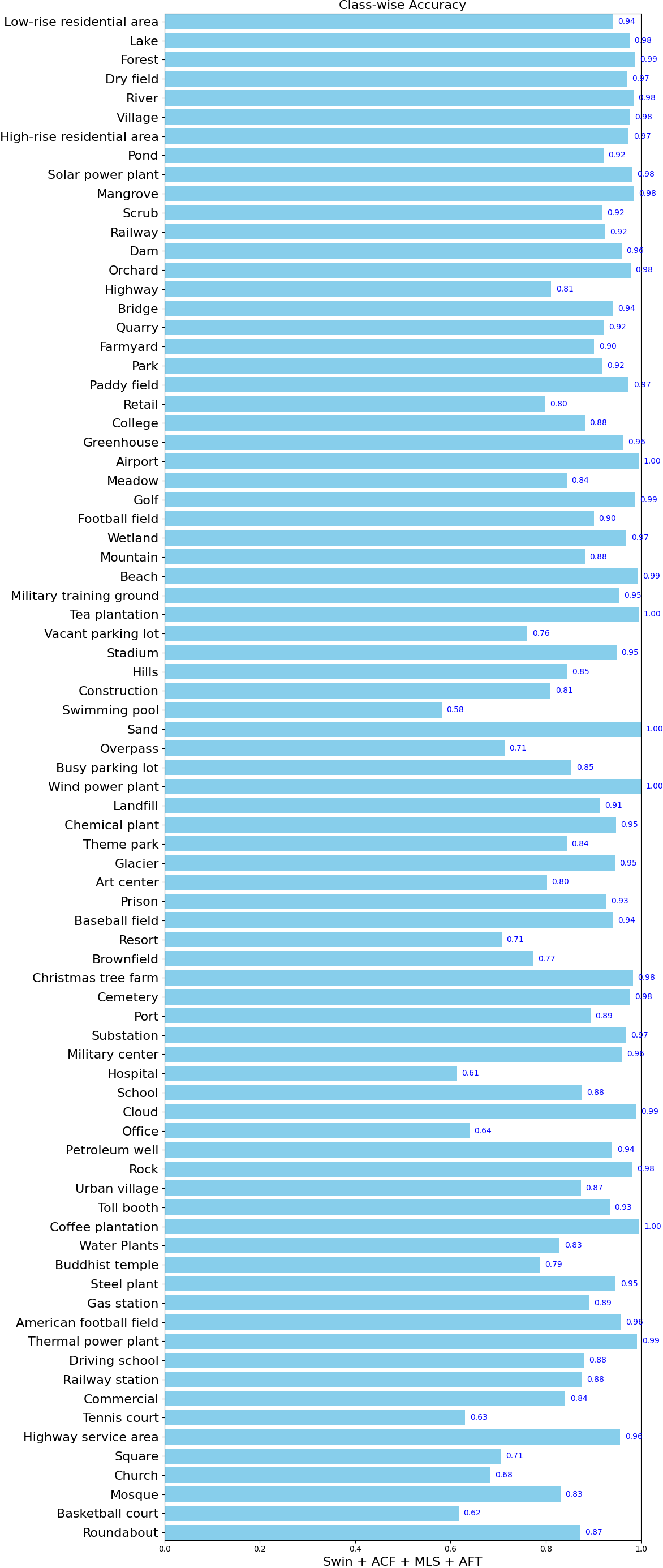}
        \label{fig:class_acc2}
    }
    \captionsetup{justification=justified, singlelinecheck=false}
    \caption{Comparison of class-wise accuracy across all 80 categories on the MEET dataset before and after applying ACF + MLS + AFT.}
    \label{fig:class_acc12}
\end{figure*}

\begin{table*}[t]
\centering
\renewcommand{\arraystretch}{1.3}
\caption{\\PERFORMANCE (\%) COMPARISON OF THE ABLATION STUDY.\\(Running Time refers to running time per one sample.)}
\label{tab:ablation}
\label{my-label}
\begin{tabular}{ccccccccccc}
\hline           
& {ACF} & {MLS} & {AFT} &  Running Time  & Parameters   & {OA} & BA\textsubscript{many}& BA\textsubscript{med} & BA\textsubscript{few}  & {BA} 
\\ \hline
& ×     & ×       & ×   & 0.0139s  &     195.12M
&92.23 & 94.89 & 73.95 & 71.24 & 75.78

\\
& \Checkmark & ×     & ×      & 0.0168s  &   226.05M

&94.29 & \textbf{97.94}	&79.00	&67.80	&77.26


\\
& \Checkmark  & \Checkmark  & ×       & 0.0168s  &    226.05M    
&\underline{95.22}& 97.01& \underline{80.64}& \underline{76.65}& \underline{81.35}

\\
& \Checkmark       & \Checkmark     & \Checkmark    & 0.0174s  &  233.15M       
&\textbf{95.87} & \underline{97.04}&\textbf{82.14}	& \textbf{80.05}	& \textbf{83.38}	

\\ \hline
\end{tabular}
\end{table*}

\begin{table*}
\centering
\caption{NUMBER OF ANNOTATED BLOCKS IN THE UFZ EVALULATION DATASET.}
\label{tab:blocks_annotation}
\begin{tabular}{lllllllllr}
\toprule
Location &  Res. & Com. & Ind. & Tra. & Edu. & Med. & Spo. & Par. & All \\ 
\midrule
Wuhan  &710&6&182&10&391&4&29&241& 1,573 \\
Shanghai &784&34&89&119&69&4&38&110& 1,247 \\
Guangzhou &196&43&107&18&155&4&165&393 & 1,081 \\
Lanzhou &95&9& 5&0&26&2&11&78 & 226 \\
Yulin &129 &2&2&6&49&4&7&7 & 196 \\
\midrule
Total & 1,914&94&385&143&690&18&250&829 & 4,323\\
\bottomrule
\end{tabular}
\end{table*}

\begin{table*}
\centering
\caption{PERFORMANCE (\%) COMPARISON OF DIFFERENT EXPERIMENTAL SETTINGS ON UFZ.}
\label{tab:my-table}
\begin{tabular}{llcccccccccc}
\toprule
Dataset  & Method & Res. & Com. & Ind. & Tra. & Edu. & Med. & Spo. & Par. & OA & BA \\ \midrule
AID & RVSA & 52.87 & 51.06 & \textbf{71.43} & 42.66 & 26.67 & 0.00 & 10.40 & 91.07& 54.61 & 43.27 \\
NWPU & MTP & 11.08 & \textbf{63.83} & 70.91 & 81.82 & 0.00 & 0.00 & 83.60 & 52.47 & 30.21 & 45.46\\
MEET & CAT & \textbf{93.73} & 59.57 & 66.49 & \textbf{88.11} & \textbf{58.26} & \textbf{50.00} & \textbf{90.00} & \textbf{90.83} & \textbf{83.76} & \textbf{74.63}

\\ \bottomrule
\end{tabular}
\end{table*}

\subsection{Results and Analysis}
The benchmark and experimental results of FGSC on the MEET dataset are shown in Table~\ref{tab:method_comparison}.
The experimental results indicate that when using Swin-Large as the backbone, our method outperforms comparison methods on the MEET dataset benchmark. Our CAT achieves an OA of 95.87\% and a BA of 83.38\%. Compared to all baselines with a similar number of parameters (excluding Swin-Huge), our method shows an improvement of nearly 1\% in OA and over 4\% in BA compared to methods using center scenes as input. Compared to methods using global scenes as input, our method shows an improvement of over 0.3\% in OA and over 1.8\% in BA. These results highlight a significant advantage across all evaluated metrics, surpassing both methods specifically designed for scene classification and those generally proposed for image recognition.
Additionally, compared to some other backbone networks~\citep{yu2024inceptionnext,sun2019deep,ding2022davit}, the Swin-Large model outperforms other methods significantly by incorporating the ACF to fully utilize contextual image information and using MLS and AFT methods to further enhance performance. Specifically, the performance gains come from the model's more powerful feature extraction capabilities. The model can incorporate complementary cues from surrounding imagery for the center scene, especially for cases where the center scene lacks prominent visual features. Additionally, the model does not overfit due to the large amount of redundant information in the surrounding imagery. This is reflected in the performance gains for the tail classes in terms of BA, as shown in Table~\ref{tab:ablation}.

To further demonstrate the effectiveness and generalization capabilities of our CAT, we also conduct experiments using the large fundation model, namely Swin-Huge from SkySense~\citep{guo2024skysense}. It can be observed that the model's performance is further enhanced when employing our CAT, achieving the best performance on both OA and BA, with scores of 97.74\% and 89.37\%, respectively, thanks to the pre-training parameter methodology of the SkySense model. 
Compared to methods using Swin-Huge as backbone, CAT shows an improvement of over 2.8\% in OA and over 9.5\% in BA. Therefore, it can be further verified that CAT achieves stable performance improvements across backbone models of different sizes.
For the following ablation study, we opt to use the Swin-Large version to reduce the cost of training resources.

As shown in Fig.~\ref{fig:case_all}, the top 3 predictions from various comparison methods are presented, as well as those provided by CAT. It can be seen that CAT not only performs the best in classification but also achieves the highest prediction confidence. This is due to CAT's strong capability in adaptive feature extraction of spatial context, ultimately leading to more stable and accurate classification results.

To illustrate the superiority of our CAT in fusing multi-level contexts, we design several classical context fusion strategies, as shown in Fig.~\ref{fig:simple}. These strategies fuse at the input, feature, and decision levels, respectively. The performance results in Table~\ref{tab:fusion} demonstrate that our CAT achieves the best performance through ACF.

Overall, comparison results on multiple metrics demonstrate that our CAT is highly effective. It makes full use of contextual information for feature extraction, achieving the best performance. Moreover, each component of our method is independent of specific scene classification methods, allowing it to seamlessly adapt and enhance performance across most proposed backbones without encountering specific limitations. This observation highlights the versatility and applicability of our proposed method.

\begin{figure*}[htbp]
    \makeatletter
    \renewcommand{\@thesubfigure}{\hskip\subfiglabelskip}
    \makeatother
    \centering
    \subfigure{
    \includegraphics[width=0.96\linewidth]{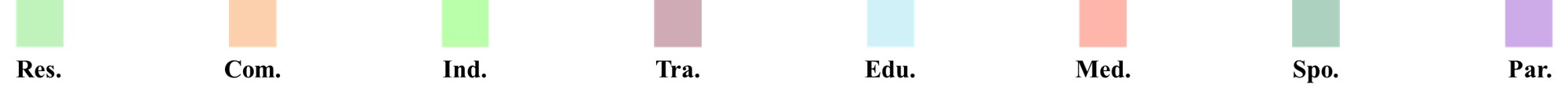}
    }
    \subfigure[(a) Ground truth]{
        \includegraphics[width=0.465\linewidth]{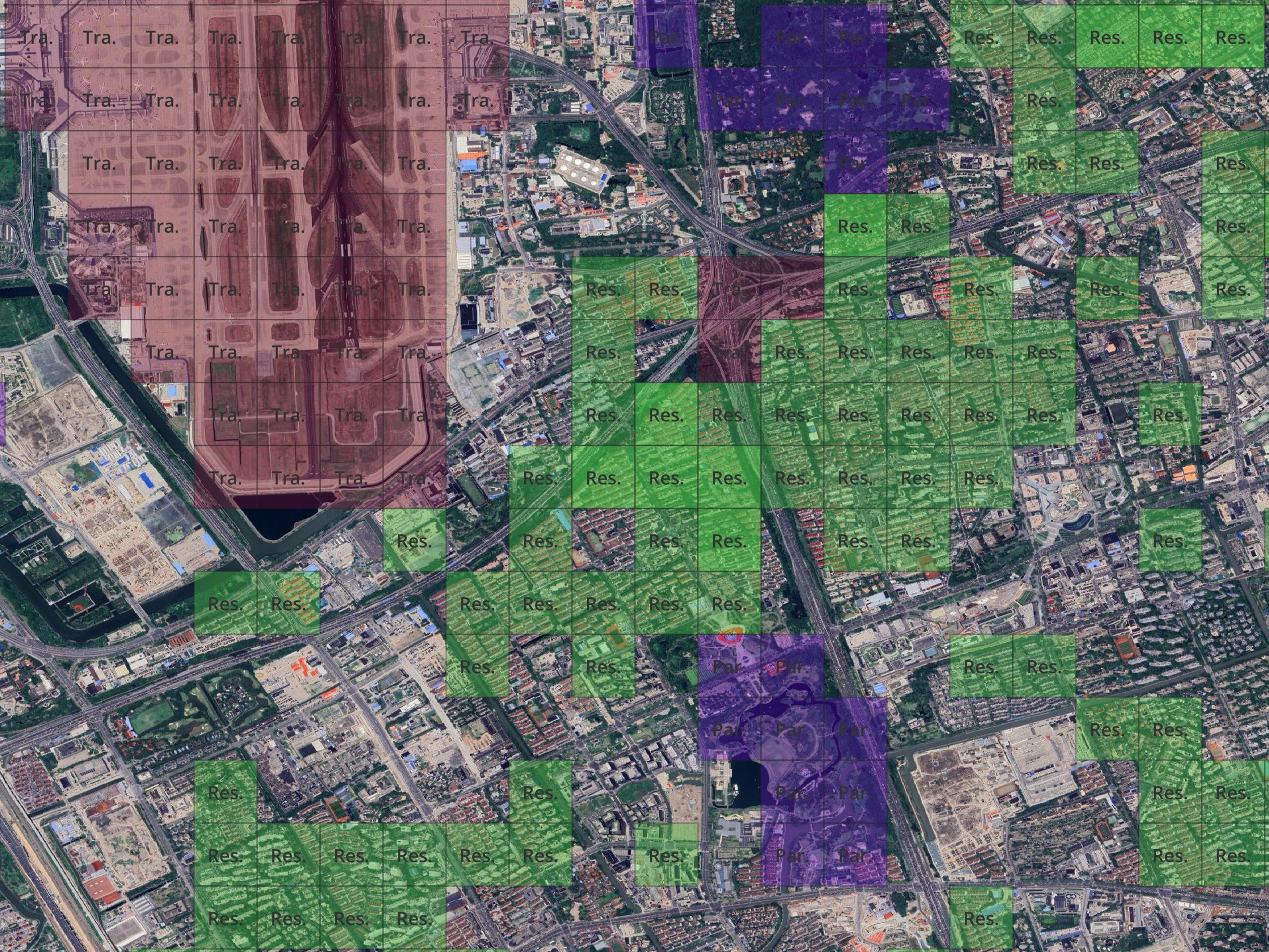}
        \label{fig:top_left_image}
    }
    \subfigure[(b) NWPU + MTP]{
        \includegraphics[width=0.465\linewidth]{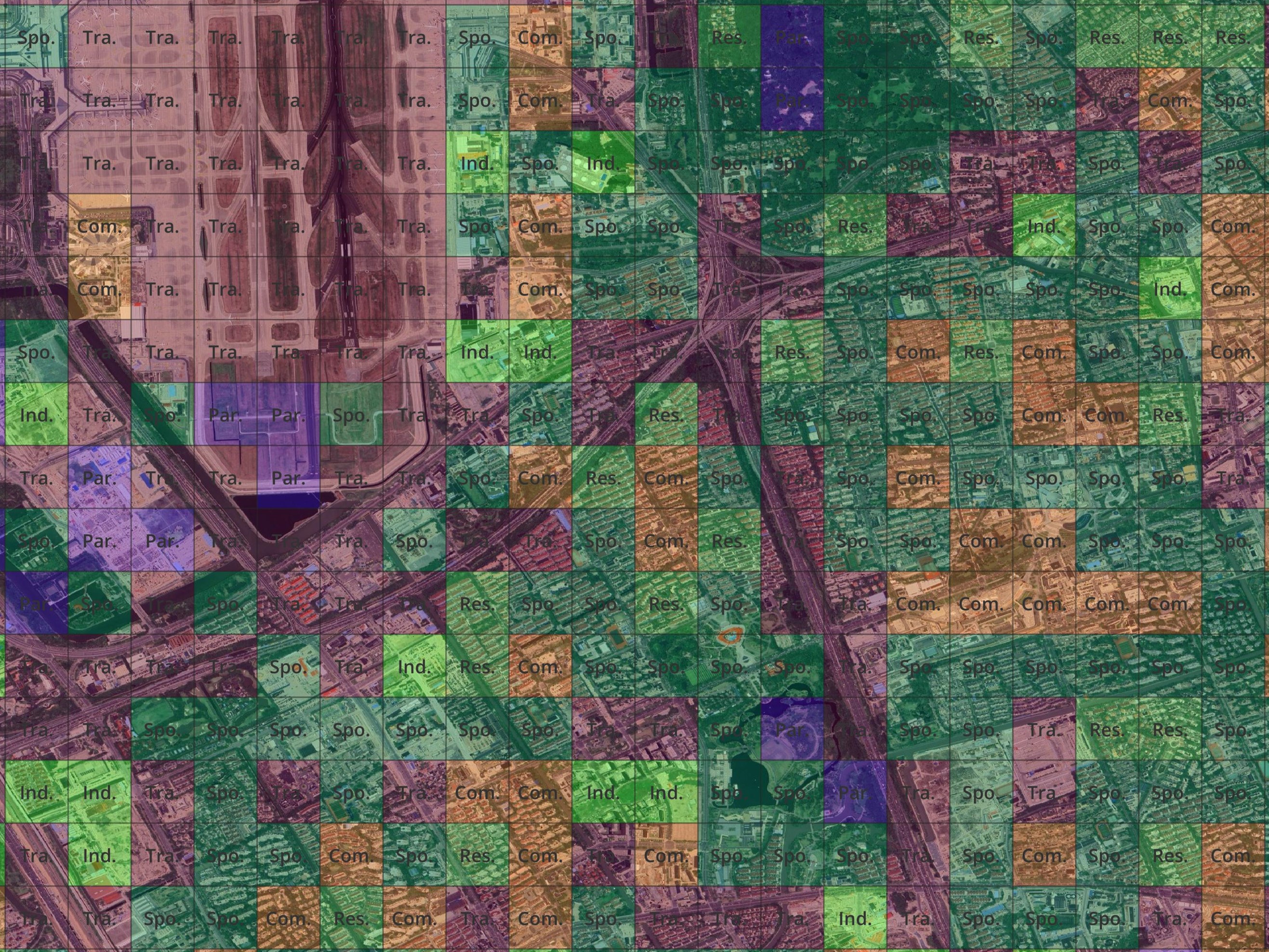}
        \label{fig:top_right_image}
    }


    \subfigure[(c) AID + RVSA]{
        \includegraphics[width=0.465\linewidth]{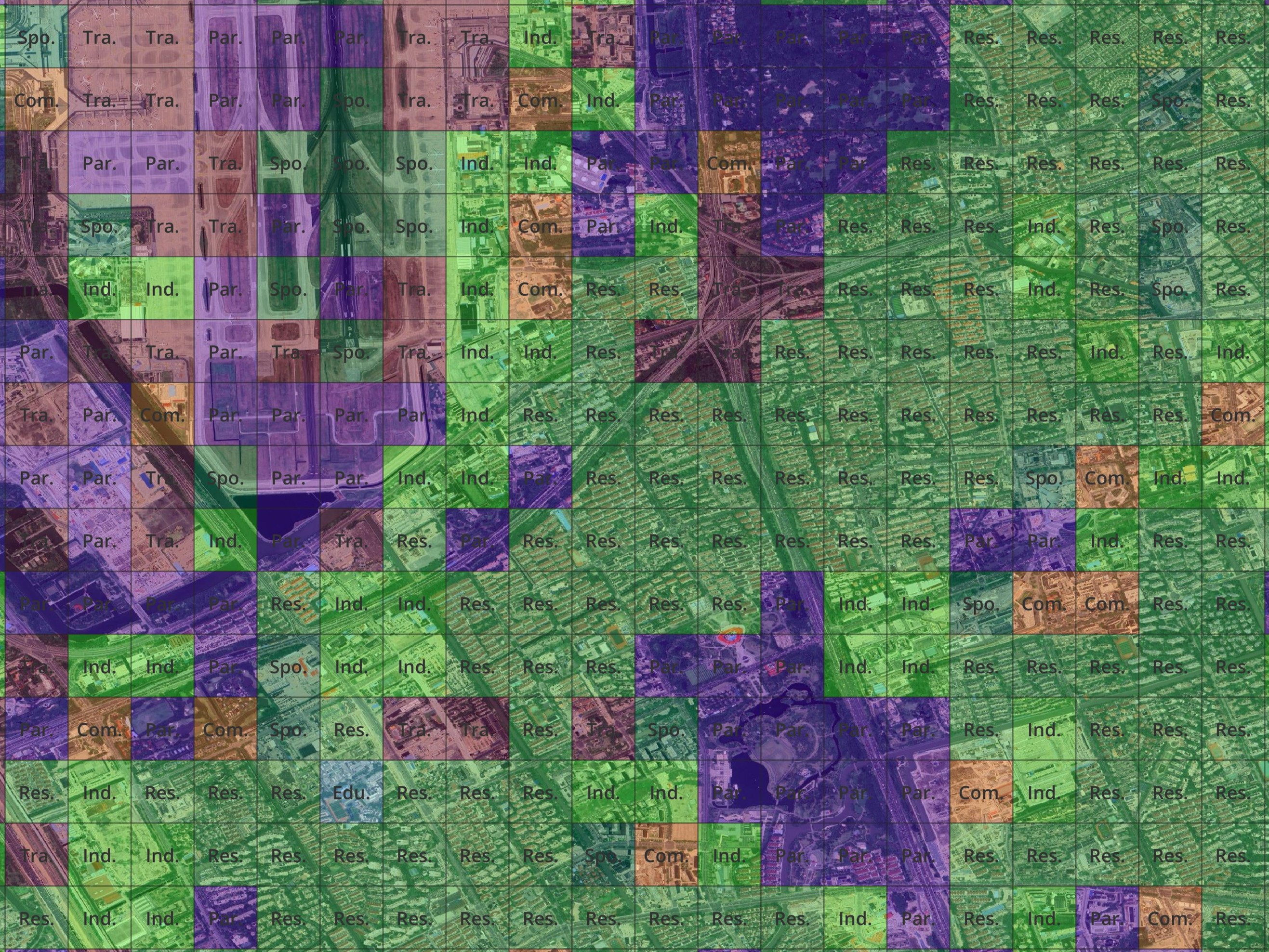}
        \label{fig:bottom_left_image}
    }
    \subfigure[(d) MEET + CAT]{
        \includegraphics[width=0.465\linewidth]{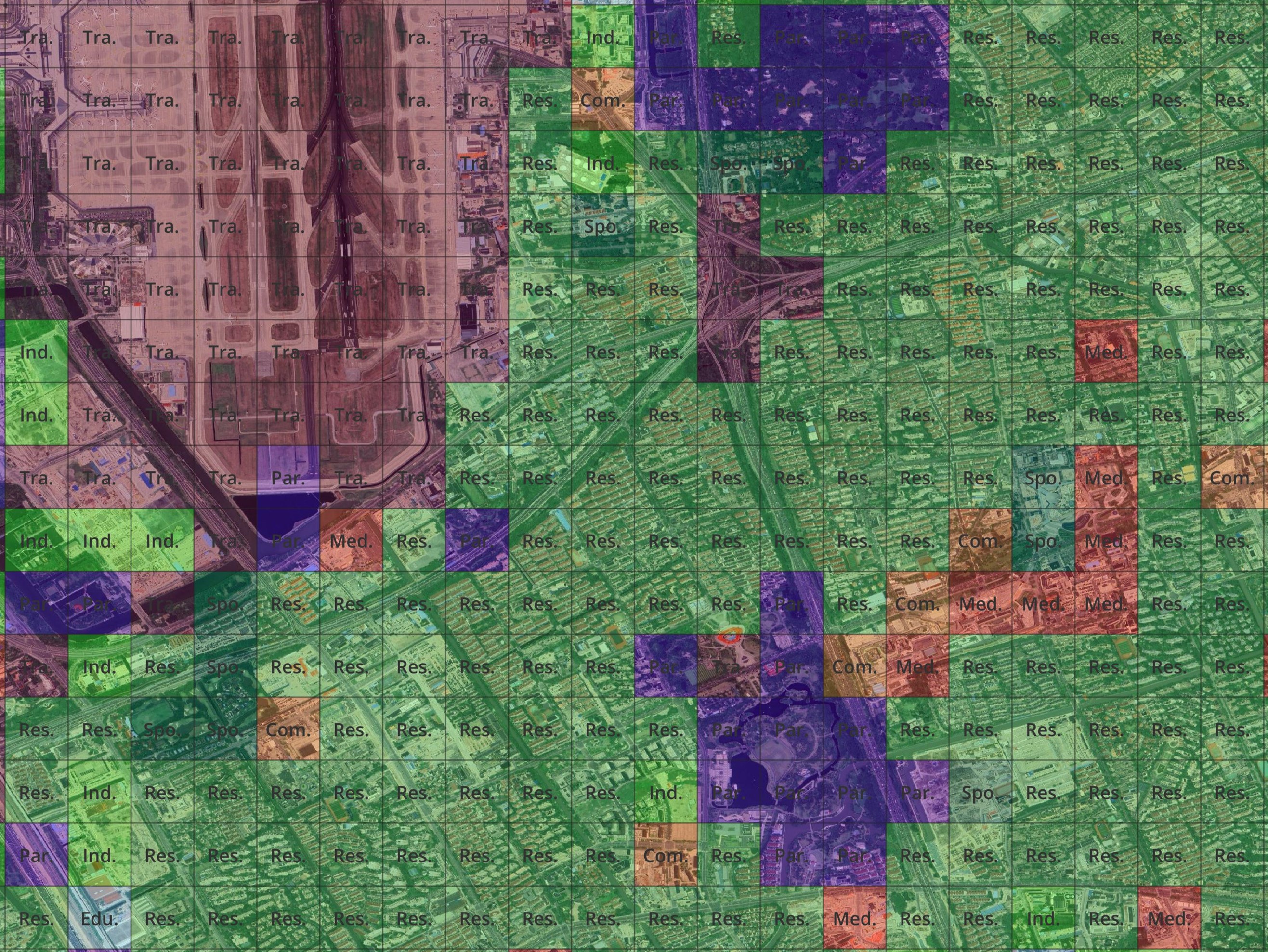}
        \label{fig:bottom_right_image}
    }
\captionsetup{justification=justified, singlelinecheck=false}
    \caption{Illustration of the mapping results of different combinations of dataset and model on the pilot area of Shanghai. The displayed image is a sub-region within the study area of Shanghai.}
    \label{fig:ufz1}
\end{figure*}

\begin{figure*}[htbp]
    \makeatletter
    \renewcommand{\@thesubfigure}{\hskip\subfiglabelskip}
    \makeatother
    \centering
    \subfigure{
    \includegraphics[width=0.96\linewidth]{figures/1003tuli.jpg}
    }
    \subfigure[(a) Ground truth]{
        \includegraphics[width=0.465\linewidth]{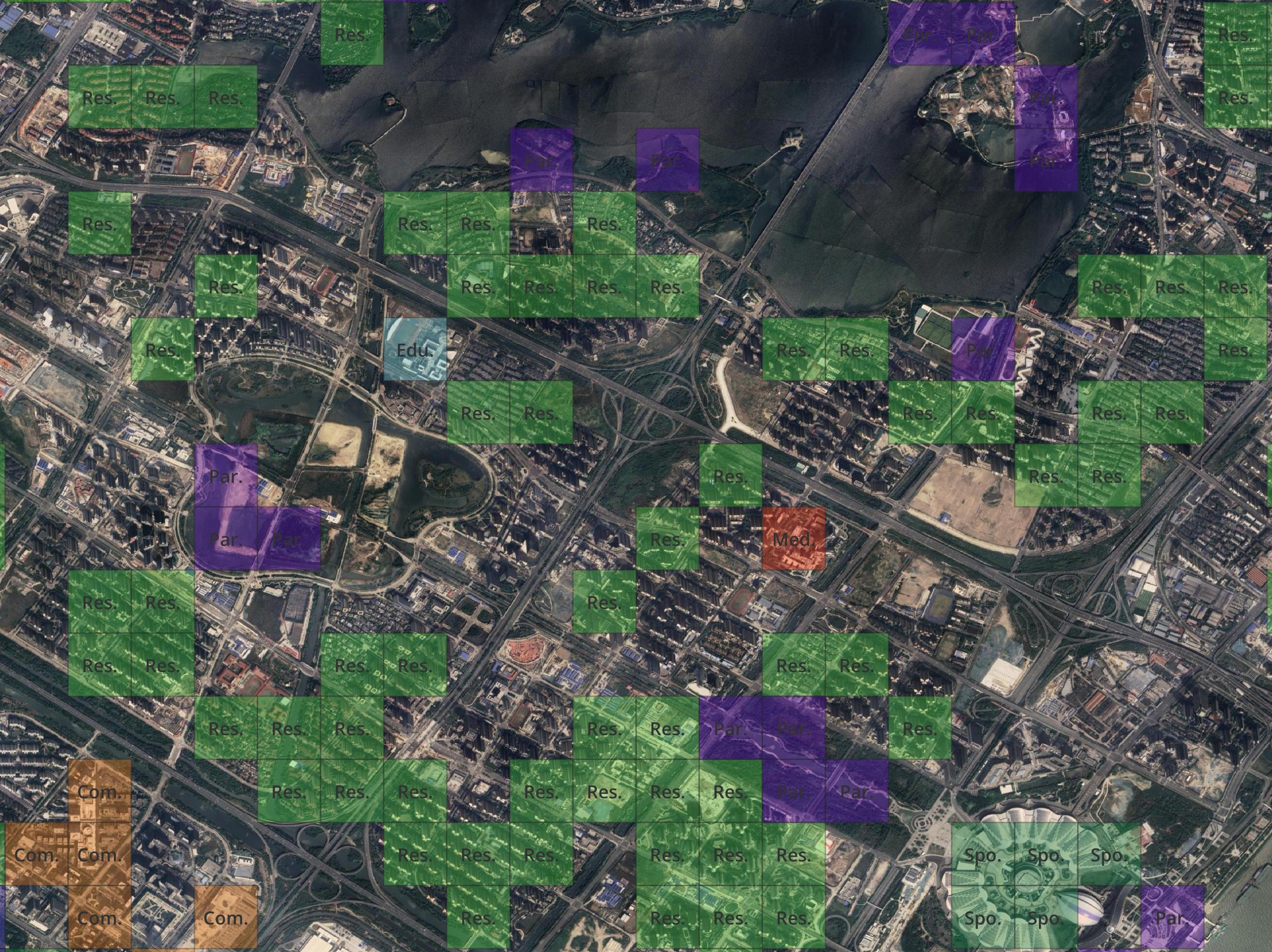}
        \label{fig:top_left_image}
    }
    \subfigure[(b) NWPU + MTP]{
        \includegraphics[width=0.465\linewidth]{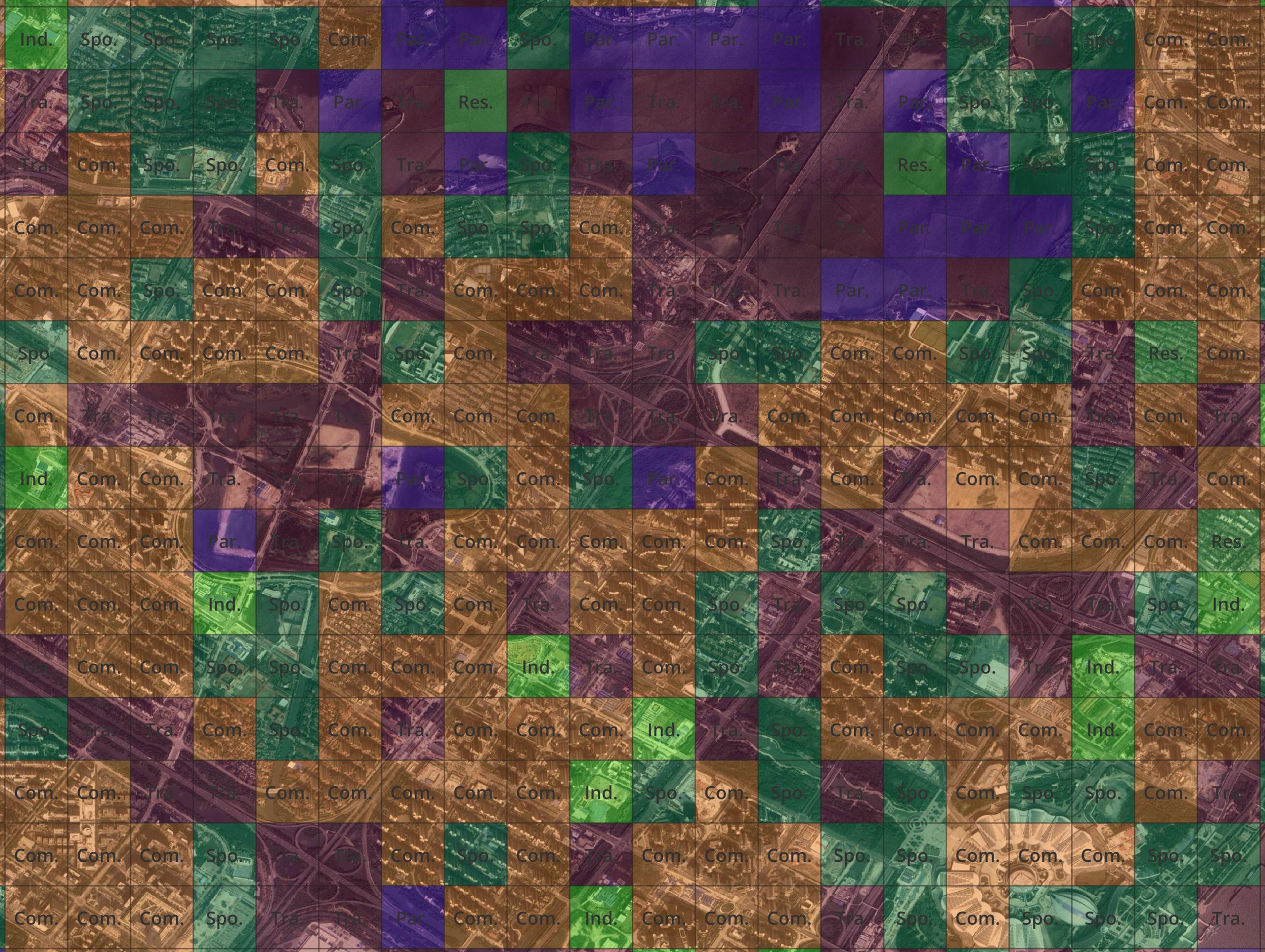}
        \label{fig:top_right_image}
    }


    \subfigure[(c) AID + RVSA]{
        \includegraphics[width=0.465\linewidth]{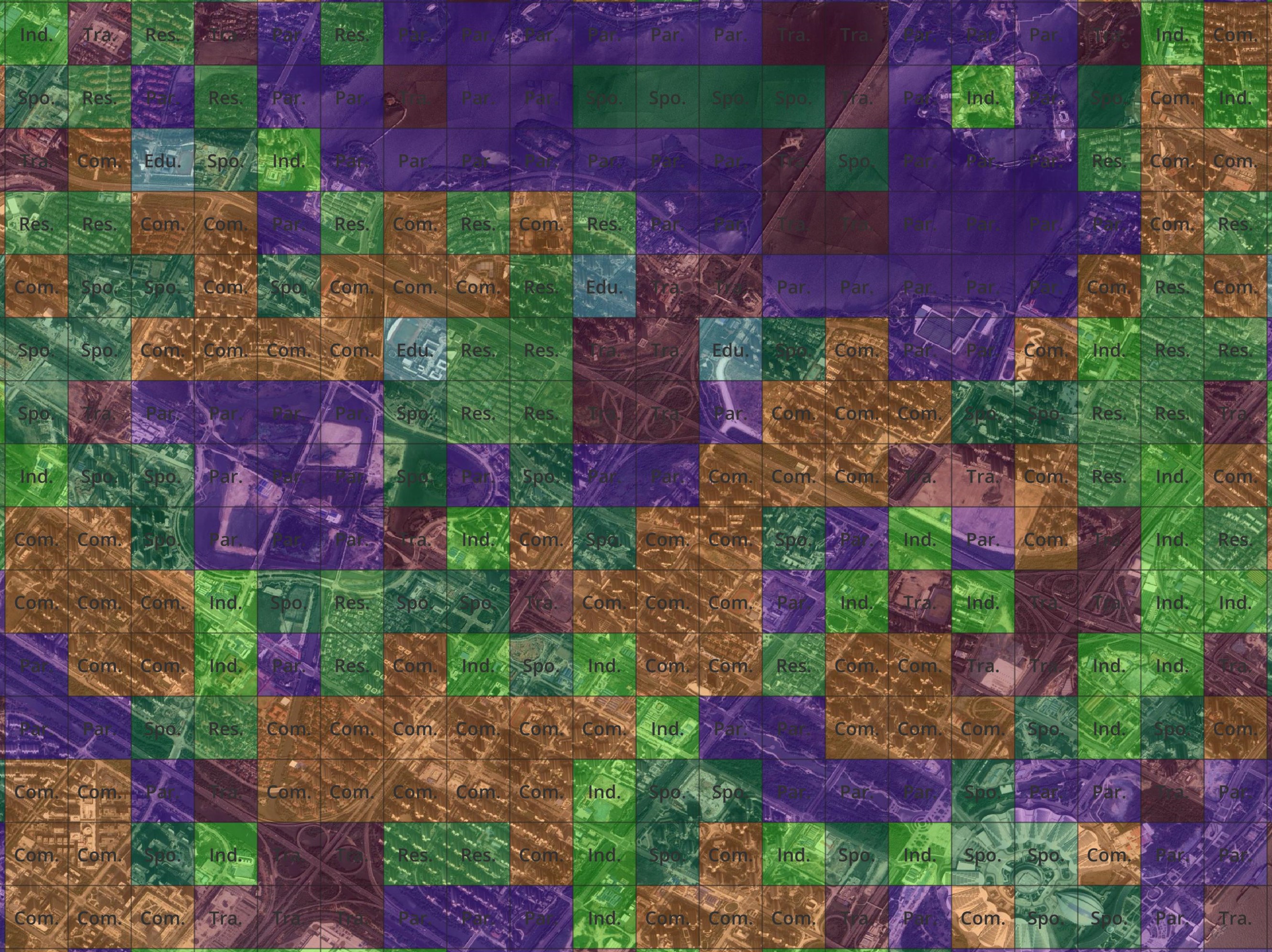}
        \label{fig:bottom_left_image}
    }
    \subfigure[(d) MEET + CAT]{
        \includegraphics[width=0.465\linewidth]{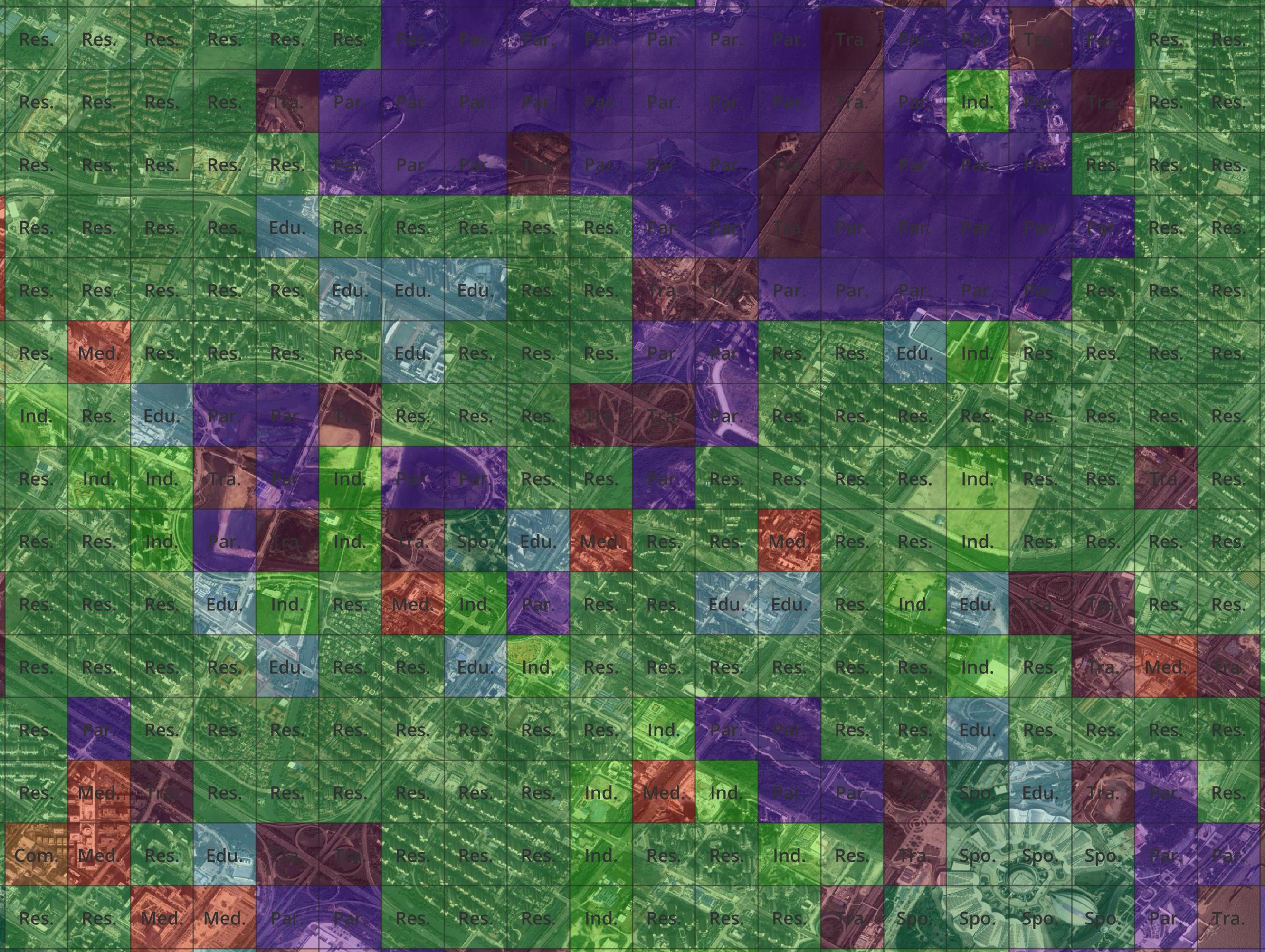}
        \label{fig:bottom_right_image}
    }
\captionsetup{justification=justified, singlelinecheck=false}
    \caption{Illustration of the mapping results of different combinations of dataset and model on the pilot area of Wuhan. The displayed image is a sub-region within the study area of Wuhan.}
    \label{fig:ufz2}
\end{figure*}

\subsection{Ablation Study of Our CAT}
The ablation experiment on the MEET dataset is to assess the impact of three key components of our proposed method: ACF, MLS and AFT.
As shown in Table~\ref{tab:ablation}, we explore the effectiveness of the proposed ACF module. The ACF module significantly improves scene classification performance by adaptively extracting high-value graphical feature information from contextual data. Compared to the baseline, OA increases by 2.06\% and BA improves by 1.48\%. This module provides the greatest performance boost by introducing incremental information and effectively merging features. 
The introduction of MLS further improves BA, indicating it effectively reduces overfitting on contextual information for most categories. Compared to using only the ACF module, MLS results in a slight decrease in head classes, due to most head classes' strong reliance on contextual information, which is different from the minority classes. The introduction of the AFT module results in increases of 0.65\% and 2.03\% in OA and BA, respectively, reflecting that we successfully enhanced the feature extraction capabilities of the three branches for different inputs without adding excessive parameters. These results highlight the capability of our method to enhance the performance of existing state-of-the-art scene classification methods. 
Additionally, we evaluate the running time per sample and the parameter number for CAT, demonstrating that each module of CAT is lightweight and does not significantly impact efficiency. To further demonstrate that CAT can mine the auxiliary image context to combat intra-class variability and inter-class similarity, we compare the class-wise accuracy across all 80 categories on the MEET dataset before and after applying CAT, as illustrated in Fig.~\ref{fig:class_acc12}. The results show a significant improvement in accuracy for the majority of categories. Specifically, the accuracy for the River category improved by 3\%, and the accuracy for the Lake category improved by 5\%.

The effectiveness of our method has been quantitatively evaluated in Table.~\ref{tab:method_comparison} and Table.~\ref{tab:ablation}. To further illustrate its capability in contextual feature extraction, we visualize the t-SNE feature map on each branch.
We employed t-SNE method to visualize the learned representative features of each ablation study setting. The perplexity for all four cases is 10 and 50 samples from all 80 geospatial scene categories are randomly selected to create a t-SNE plot. 
As shown in Fig.~\ref{fig:tsne}, after applying ACF, the embeddings do not become significantly more separable in the feature space. This may be due to the model overfitting to some extent on the auxiliary scene. However, after applying MLS to enhance the feature extraction capabilities for both the center scene and surrounding scenes, and applying the AFT module to enhance the feature capabilities of the three branches, the embeddings become significantly more separable. These results indicate that our CAT achieves better class separability at the feature level.

\subsection{Superiority Analysis of Our CAT}

In Fig.~\ref{fig:cam}, we present some examples with scores and class activation map (CAM) on each branch. The scores represent the prediction on the ground truth category (shown on the left side) after applying softmax from classification head on that branch. Changes in the scores reflect the gain in performance due to the accumulation of multi-level context.
It can be observed that the predictions consistently improve with the introduction of multi-level auxiliary scenes, which is reasonable for cases that require auxiliary scenes. However, for cases that can achieve sufficient saliency without auxiliary scenes, such as airport example in Fig.~\ref{fig:cam}, the model performance has not degraded with the introduction of redundant information. This demonstrates that our model has an adaptive context fusion capability, showing strong generalization.
With the introduction of auxiliary scenes, the model can extract more visual features from the context to interpret the center scene. Specifically, from the examples of river and lake, it can be observed that using only the center scene as input is not sufficient. After introducing auxiliary scenes, the input data includes contextual information, such as riverbanks, enabling the model to correctly differentiate the water body into a river or lake. For the village example illustrated, the visual features of fields included in the auxiliary scenes contribute to distinguish the village category from other similar categories in the center scene, such as low-rise residential area category.

\subsection{Application Evaluation of Our MEET on Urban Functional Zone Mapping}
To validate the the setting superiority of the zoom-free characteristic and scene-in-scene sample layout of our MEET dataset, we conduct experiments on urban functional zone mapping (UFZ). In the pilot application, UFZ aims to predict the land-use category of each fixed-resolution RSI block and considers 8 land-use categories:
the Residential (Res.) category denotes various types of residential buildings of different heights; the Commercial (Com.) category indicates commercial and business activities including offices, retail and malls; the Industrial (Ind.) category denotes the land with industrial purposes; the Transportation (Tra.) category include various transportation facilities; the Educational (Edu.) category denotes educational institutions including universities, colleges and primary and secondary schools; the Medical (Med.) category is primarily dedicated to healthcare facilities; the Sport and cultural (Spo.) category indicates sports and cultural activities including sports fields and art centers; the Park and greenspace (Par.) category consists of of parks, forests and other public green spaces. Across five cities (e.g., Shanghai, Lanzhou, Wuhan, Guangzhou and Yulin), we creat a UFZ evaluation dataset where one large region in each city is selected and its corresponding 1-meter spatial resolution RSI is split into blocks with 256$\times$256 pixels. A total of 4,323 blocks are manually annotated by experts in remote sensing with the above 8-class land-use classification system. To avoid ambiguous annotation, we only choose semantically clear blocks for manual labeling, which results in a relatively sparse annotation distribution. The specific annotation information is summarized in Table~\ref{tab:blocks_annotation}.



To verify the practicability and superiority of the given MEET dataset, we train models on MEET and the other widely adopted datasets such as AID and NWPU to address UFZ. As far as models, RVSA~\citep{wang2022advancing} and MTP~\citep{wang2024mtp} are selected as they are the state-of-the-art models on AID and NWPU, respectively. During inference, both the RVSA model trained on AID and the MTP model trained on NWPU adopt the center block as input for classification. To facilitate the unified evaluation, we map the classification results of models trained on AID, NWPU and MEET into the 8-class land-use classification system.

The evaluation results of different models trained on AID, NWPU, and MEET are shown in Table~\ref{tab:my-table}. The experimental results indicate that the combination of MEET and CAT achieves the best performance in most categories, with a significant improvement over the other combinations. This improvement comes from both the dataset and the method. From the dataset perspective, MEET has more fine-grained categories, enabling the model to extract more detailed semantic information from complex urban environments, leading to a more comprehensive understanding of different UFZ categories. From the method perspective, CAT can adaptively integrate auxiliary scenes, providing more stable classification performance for cases where the center scene is not very salient.
Furthermore, Fig.~\ref{fig:ufz1} and~\ref{fig:ufz2} illustrate the mapping results under various settings from two different cities. It is evident that our CAT, when trained on the MEET dataset, demonstrates a substantial performance improvement in mapping accuracy and geographical coherence. It is noted that our CAT can effectively utilize contextual information to improve classification performance even in areas with low saliency. In Fig.~\ref{fig:ufz1}, the CAT with MEET shows better classification capabilities for objects such as parks, airports, and residential buildings, thanks to the effective use of contextual information. In Fig.~\ref{fig:ufz2}, the CAT with MEET demonstrates significantly better classification performance for the low-salient categories of Educational and Medical, thanks to the MEET dataset's more comprehensive and rich subclass samples for UFZ categories. As a whole, these comparisons underscore the effectiveness of the zoom-free scene-in-scene sample layout in MEET.

\section{CONCLUSION}
\label{sec:conclusion}
In this paper, we introduce a large dataset named MEET for FGSC with zoom-free RSI. MEET is comprised of over 1.03 million samples with scene-in-scene layout, encompassing 80 fine-grained geospatial scene categories. Samples are collected globally and include multi-level spatial context information. The large sample size, the granularity of categories, and the inclusion of spatial context imagery make MEET a valuable dataset. It provides essential data conditions for advancing a challenging yet meaningful new task, FGSC with zoom-free RSI. Additionally, we propose a CAT for FGSC, which effectively integrates contextual information and achieves progressive visual feature extraction. 
Compared with existing state-of-the-art algorithms, our
CAT performs excellently on the MEET dataset, demonstrating the CAT's value from both quantitative and qualitative perspectives.

In future work, we plan to further enrich the MEET dataset in terms of sample volume and category diversity and propose global-scale scene classification mapping products. These will be made available to a broader scientific community in need of related analytical data, thereby continuously driving progress in this research area.



%


%

\ifCLASSOPTIONcaptionsoff
  \newpage
\fi



\normalem
\bibliographystyle{IEEEtran}
\bibliography{custom}
%

%





\begin{IEEEbiography}[{\includegraphics[height=1.25in,clip,keepaspectratio]{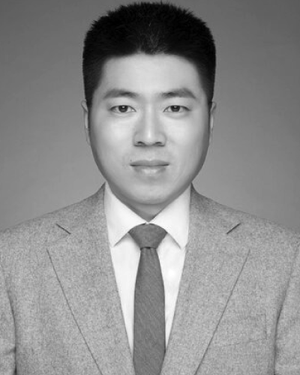}}]{Yansheng Li}
(Senior Member, IEEE) received the B.S. degree in information and computing science from Shandong University, Weihai, China, in 2010, and the Ph.D. degree in pattern recognition and intelligent system from the Huazhong University of Science and Technology, Wuhan, China, in 2015.  He is currently a Full Professor and Vice Dean with the School of Remote Sensing and Information Engineering, Wuhan University, Wuhan, China. He has authored more than 100 peer-reviewed papers such as IEEE TPAMI, RSE, IEEE TIP, CVPR, ECCV and AAAI. His research interests include knowledge graph, deep learning and their applications in remote sensing big data mining. He is an Associate Editor of IEEE TGRS and a Junior Editorial Member of The Innovation.
\end{IEEEbiography}
\begin{IEEEbiography}[{\includegraphics[height=1.25in,keepaspectratio]{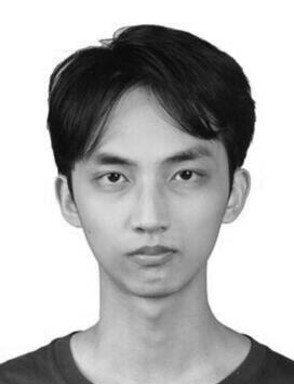}}]{Yuning Wu}
received the B.S. degree in computer science and technology from Wuhan University, Wuhan, China in 2022. He is currently pursuing his M.S. degree with the School of Remote Sensing and Information Engineering, Wuhan University, Wuhan, China. His research interests include remote sensing scene classification and few-shot learning.
\end{IEEEbiography}

\begin{IEEEbiography}[{\includegraphics[height=1.25in,keepaspectratio]{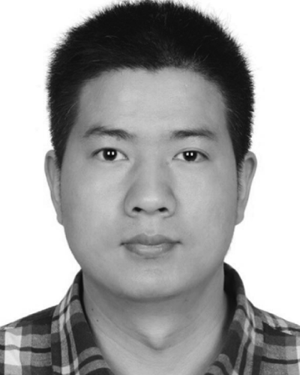}}]{Gong Cheng}
(Member, IEEE) received the B.S. degree in biomedical engineering from Xidian University, Xi’an, China, in 2007, and the M.S. and Ph.D. degrees in pattern recognition and intelligent systems from Northwestern Polytechnical University, Xi’an, China, in 2010 and 2013, respectively.,He is currently a Professor with Northwestern Polytechnical University, Xi’an, China. His research interests include computer vision and pattern recognition.
\end{IEEEbiography}

\begin{IEEEbiography}[{\includegraphics[height=1.25in,keepaspectratio]{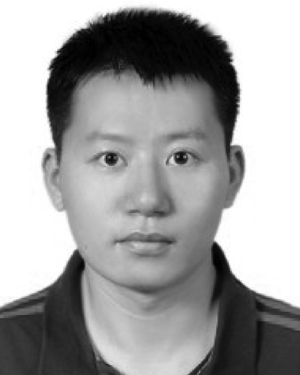}}]{Chao Tao}
received the B.S. degree from the School of Mathematics and Statistics, Huazhong University of Science and Technology, Wuhan, China, in 2007, and the Ph.D. degree from the Institute for Pattern Recognition and Artificial Intelligence, Huazhong University of Science and Technology, in 2012. He is currently an Associate Processor with the School of Geosciences and Info-Physics, Central South University, Changsha, China. He has authored more than 30 peer-reviewed articles in international journals from multiple domains, such as remote sensing and computer vision. His research interests include computer vision, machine learning, deep learning, and their applications in remote sensing. He has been frequently serving as a Reviewer for more than four international journals, including the IEEE Transactions on Geoscience and Remote Sensing (IEEE-TGRS), IEEE Geoscience and Remote Sensing Letters (IEEE-GRSL), IEEE Journal of Selected Topics in Applied Earth Observations and Remote Sensing (IEEE-JSTAR), PERS, and ISPRS. He is also a Communication Evaluation Expert for the National Natural Science Foundation of China.
\end{IEEEbiography}

\begin{IEEEbiography}[{\includegraphics[height=1.25in,keepaspectratio]{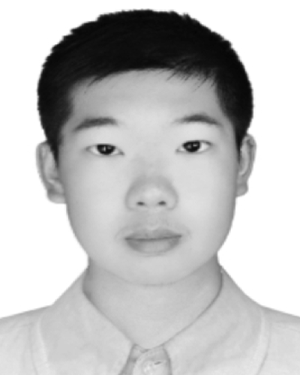}}]{Bo Dang}
received the B.S. degree in remote sensing science and technology from Wuhan University, Wuhan, China in 2022. He is currently working toward the Ph.D. degree with the School of Remote Sensing and Information Engineering, Wuhan University. He has published several papers in CVPR, AAAI, ISPRS Journal of Photogrammetry and Remote Sensing, etc. His research interests include remote sensing semantic segmentation and remote sensing foundation model.
\end{IEEEbiography}

\begin{IEEEbiography}[{\includegraphics[height=1.25in,keepaspectratio]{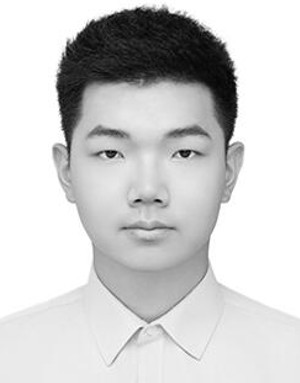}}]{Yu Wang}
received the B.S. degree in School of Remote Sensing and Information Engineering, Wuhan University, Wuhan, China in 2023. He is currently pursuing his M.S. degree with the School of Remote Sensing and Information Engineering, Wuhan University, Wuhan, China. His research interests include deep learning and knowledge graph.
\end{IEEEbiography}

\begin{IEEEbiography}[{\includegraphics[height=1.25in,keepaspectratio]{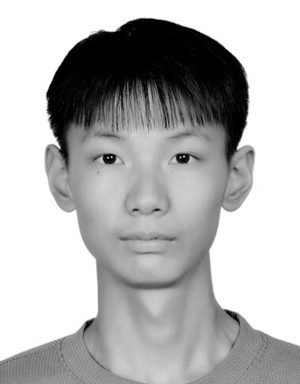}}]{Jiahao Zhang}
is currently pursuing his B.S. degree with the School of Remote Sensing and Information Engineering, Wuhan University, Wuhan, China. His research focuses on developing unified modeling frameworks from multi-source geospatial data.
\end{IEEEbiography}

\begin{IEEEbiography}[{\includegraphics[height=1.25in,keepaspectratio]{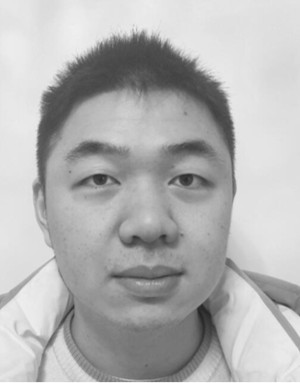}}]{Chuge Zhang}
is currently pursuing his B.S. degree with the School of Remote Sensing and Information Engineering, Wuhan University, Wuhan, China. His research interests include remote sensing image segmentation and model compression.
\end{IEEEbiography}

\begin{IEEEbiography}[{\includegraphics[height=1.25in,keepaspectratio]{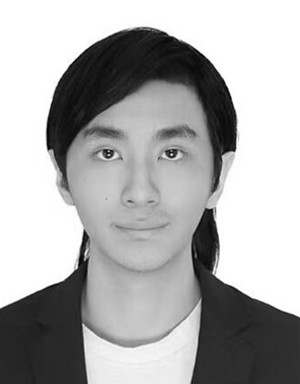}}]{Yiting Liu}
is currently pursuing his B.S. degree with the School of Remote Sensing and Information Engineering, Wuhan University, Wuhan, China. His research interests include remote sensing scene classification.
\end{IEEEbiography}

\begin{IEEEbiography}[{\includegraphics[height=1.25in,keepaspectratio]{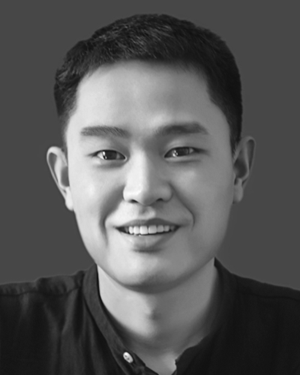}}]{Xu Tang}
(Senior Member, IEEE) received the B.S., M.S., and Ph.D. degrees in electronic circuit and system from Xidian University, Xi'an, China, in 2007, 2010, and 2017, respectively.,From 2015 to 2016, he was a Joint Ph.D. Student along with Prof. W. J. Emery with the University of Colorado at Boulder, Boulder, CO, USA. He is currently an Associate Professor with the Key Laboratory of Intelligent Perception and Image Understanding, Ministry of Education, Xidian University. His research interests include remote sensing image content-based retrieval and reranking, hyperspectral image processing, remote sensing scene classification, and object detection.
\end{IEEEbiography}

\begin{IEEEbiography}[{\includegraphics[height=1.25in,keepaspectratio]{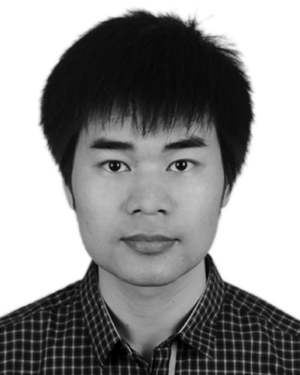}}]{Jiayi Ma}
(Senior Member, IEEE) received the B.S. degree in information and computing science and the Ph.D. degree in control science and engineering from the Huazhong University of Science and Technology, Wuhan, China, in 2008 and 2014, respectively.,He is currently a Professor with Electronic Information School, Wuhan University, Wuhan, China. He has authored or coauthored more than 200 refereed journals and conference papers, including IEEE Transactions on Pattern Analysis and Machine Intelligence (TPAMI)/IEEE Transactions on Image Processing (TIP), International Journal of Computer Vision (IJCV), Computer Vision and Pattern Recognition Conference (CVPR), International Conference on Computer Vision, and European Conference on Computer Vision. His research interests include computer vision, machine learning, and pattern recognition.,Dr. Ma has been identified in the 2019–2022 Highly Cited Researcher lists from the Web of Science Group. He is an Area Editor of Information Fusion and an Associate Editor of Neurocomputing, Sensors, and Entropy.
\end{IEEEbiography}

\begin{IEEEbiography}[{\includegraphics[height=1.25in,keepaspectratio]{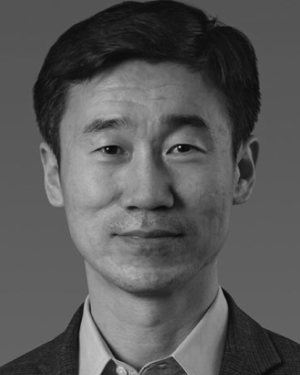}}]{Yongjun Zhang}
(Member, IEEE) received the B.S. degree in geodesy, the M.S. degree in geodesy and surveying engineering, and the Ph.D. degree in geodesy and photography from Wuhan University, Wuhan, China, in 1997, 2000, and 2002, respectively. He is currently a full professor and dean with the School of Remote Sensing and Information Engineering, Wuhan University. He has published more than 150 research articles and one book. His research interests include aerospace and low-attitude photogrammetry, image matching, combined block adjustment with multisource datasets, artificial intelligence-driven remote sensing image interpretation, and 3-D city reconstruction.
\end{IEEEbiography}

\end{document}